\documentclass[letterpaper, 10pt, journal, twoside]{IEEEtran}

\usepackage[left=0.6in,right=0.6in,top=0.75in,bottom=0.75in]{geometry}

\usepackage{times}
\usepackage[hidelinks]{hyperref}
\usepackage{authblk}

\usepackage{cite}
\usepackage{amsmath,amssymb,amsfonts,amsthm}
\usepackage{algorithmic}
\usepackage{graphicx}
\usepackage{algorithm,algorithmic}
\usepackage{hyperref}
\hypersetup{hidelinks}
\usepackage{textcomp}
\usepackage{color}

\usepackage{booktabs}
\usepackage{multirow}

\usepackage{mathtools}
\usepackage[capitalize]{cleveref}

\mathtoolsset{showonlyrefs=false}

\newtheorem{assumption}{Assumption}[section]
\newtheorem{lemma}{Lemma}
\newtheorem{theorem}{Theorem}
\newtheorem{corollary}{Corollary}

\newenvironment{theorem*}[1]
  {\par\addvspace{\topsep}\noindent{\bfseries #1.}\enspace\itshape}
  {\par\addvspace{\topsep}}

\newcommand\numberthis{\addtocounter{equation}{1}\tag{\theequation}}

\newcommand{\dwass}{\mathcal{W}_1}

\crefname{assumption}{assumption}{assumptions}
\Crefname{assumption}{Assumption}{Assumptions}
\crefname{section}{section}{sections}
\Crefname{section}{Section}{Sections}

\usepackage{comment}
\newif\ifarxiv         \arxivtrue
\newif\ifmarkrevisions \markrevisionsfalse

\ifarxiv
  \long\def\revised#1{#1}
  \long\def\postrevision#1{#1}
  
  \excludecomment{journalonly}
\else
  \definecolor{draftblue}{rgb}{0.05,0.33,0.62}
  \definecolor{postrevisionpurple}{rgb}{0.48,0.16,0.58}
  \ifmarkrevisions
    \long\def\revised#1{{\color{draftblue}#1}}
    \long\def\postrevision#1{{\color{postrevisionpurple}#1}}
  \else
    \long\def\revised#1{#1}
    \long\def\postrevision#1{#1}
  \fi
  
  \excludecomment{arxivonly}
\fi

\def\BibTeX{{\rm B\kern-.05em{\sc i\kern-.025em b}\kern-.08em
T\kern-.1667em\lower.7ex\hbox{E}\kern-.125emX}}
\title{Nonasymptotic CLT and Error Bounds for Linear Two-Time-Scale Stochastic Approximation}
\author{Seo Taek Kong, Sihan Zeng, Thinh T. Doan, and R. Srikant
\thanks{This work was supported in part by the National Science Foundation (NSF) under Grant CNS 23-12714, CCF 22-07547, CAREER 2339509, CCF 2527044, and AFOSR Grant FA9550-24-1-0002.}
\thanks{Seo Taek Kong and R. Srikant are with the Coordinated Science Laboratory, Department of Electrical and Computer Engineering, University of Illinois Urbana-Champaign, Urbana, IL 61801 USA (e-mail: skong10@illinois.edu; rsrikant@illinois.edu).}%
\thanks{Sihan Zeng is with JP Morgan AI Research, Palo Alto, CA 94304 USA (e-mail: szeng2017@gmail.com).}%
\thanks{Thinh T. Doan is with the Aerospace Engineering and Engineering Mechanics Department at University of Texas, Austin, Austin, TX 78712 USA (e-mail: thinhdoan@utexas.edu).}%
}

\begin{document}

\maketitle

\begin{abstract}
    We consider linear two-time-scale stochastic approximation algorithms driven by martingale noise.
    Recent applications in machine learning motivate the need to understand finite-time error rates, but conventional stochastic approximation analyses focus on either asymptotic convergence in distribution or finite-time bounds that are far from optimal.
    Prior work on asymptotic central limit theorems (CLTs) suggests that two-time-scale algorithms may be able to achieve $1/\sqrt{K}$ error in expectation, with a constant given by the expected norm of the limiting Gaussian vector.
    However, the best known finite-time rates are much slower.
    We derive the first nonasymptotic Wasserstein-1 CLT \revised{for linear two-time-scale stochastic approximation with Polyak-Ruppert averaging driven by martingale difference noise}.
    As a corollary, we show that the expected error achieved by Polyak-Ruppert averaging decays at rate $1/\sqrt{K}$, which significantly improves on the rates of convergence in prior works.
\end{abstract}


\section{Introduction}\label{sec:introduction}
\IEEEPARstart{S}{tochastic} approximation (SA) was introduced by \cite{RobbinsMonro} as an iterative sample-based approach to find the root (or fixed point)  $x^{\star}$  of some unknown operator $F$.
In particular, SA iteratively updates an estimate $x$ by moving along the direction of the sample $F(x;\xi)$ scaled by some step size, where $\xi$ is a random variable representing sampling noise.
The simplicity of its implementation has enabled broad applications in many areas including stochastic optimization, machine learning, and reinforcement learning \cite{SBbook2018,LanBook2020}.
The convergence properties of SA are well-studied, where the error rate achieved by SA with Polyak-Ruppert (PR) averaging is known to decay faster than that of SA without PR averaging under widely-studied conditions \cite{polyakJuditsky,lauand2024revisiting}.

In this paper, we analyze the finite-time error rates of linear two-time-scale SA (TSA) algorithms, a generalized variant of SA, used to solve a system of two coupled linear equations.
The goal of TSA is to find a pair $(x^{*},y^{*})$ that solves
\begin{align}\label{eq:problem}
\left\{\begin{array}{ll}
A_{ff}x^{*}+A_{fs}y^{*} = \vartheta_f,\\
A_{sf}x^{*}+A_{ss}y^{*} = \vartheta_s,
\end{array}\right.
\end{align}
\revised{where $\vartheta_f, \vartheta_s$ are unknown (deterministic) right-hand-side vectors} and $A_{ff}, A_{fs}, A_{sf}, A_{ss}$ are the system parameters.
In this setting, TSA iteratively updates $(x,y)$ as follows:
\begin{equation}
    \begin{split}
        x_{k+1} &= x_k - \alpha_k \left(A_{ff} x_k + A_{fs} y_k - \vartheta_f - W_k \right) \\
        y_{k+1} &= y_k - \gamma_k \left(A_{sf} x_k + A_{ss} y_k - \vartheta_s - V_k \right),
    \end{split}
    \label{eq:ttsa}
\end{equation}
where $\{(W_k, V_k)\}$ is a martingale difference sequence representing sampling noise.
Here, $\alpha_{k} > \gamma_{k}$ are two different step sizes. The variable $x$ is often referred to as the fast-time-scale variable (using larger step sizes) while $y$ is called the slow-time-scale variable (using smaller step sizes).

TSA has received a great amount of interest due to its applications where classic SA is not applicable.
Prominent examples include reinforcement learning (e.g., gradient temporal-difference learning \cite{sutton2008convergent,sutton2009fast,xu2019two} and actor-critic methods \cite{konda1999actor,bhatnagar2012stochastic,harsh19,wu2020finite,zeng2024accelerated}) and min-max optimization \cite{zeng2022regularized}.
These methods require one iterate to be updated using a smaller step size than the other to guarantee convergence.

We use TSA to refer to the sequence $\{(x_k, y_k)\}$ in \eqref{eq:ttsa} without averaging to distinguish from TSA-PR defined as $\bar{x}_K = K^{-1} \sum_{k=1}^K x_k$ and $\bar{y}_K = K^{-1} \sum_{k=1}^K y_k$.
Polyak-Ruppert averaging is known to reduce the asymptotic covariance (or mean square error) for single-time-scale algorithms \cite{polyakJuditsky} and for two-time-scale algorithms \cite{mokkadem2006convergence}.
Finite-time error bounds, inspired by applications in machine learning, have been established only recently for SA \cite{srikant19} and TSA~\cite{dalal2018finite,kaledin2020finite,haque2023tightfinitetimebounds} without PR averaging.
Finite-time error bounds are useful for many reasons including algorithm design, where step sizes can be chosen to minimize the error bounds, or understanding when to halt an algorithm to achieve a desired error tolerance.
More recently, finite-time bounds have been established for SA-PR in the form of non-asymptotic central limit theorems (CLTs) \cite{srikant2024CLT,samsonov2024gaussian}.
\revised{A non-asymptotic CLT in the Wasserstein-1 metric provides a rich characterization: it proves exact convergence of the expected error to its asymptotic limit (as in \Cref{cor:mae}) and characterizes the full distributional shape, enabling probability approximations over convex sets \cite{nourdin2022multivariate}.}
We build upon this line of work by establishing the first nonasymptotic Wasserstein-1 CLT for linear TSA-PR driven by martingale difference noise.

\noindent \textbf{Our contributions} are summarized below.
\begin{enumerate}
    \item We establish in \Cref{lem:mse} the convergence of second moments for TSA (without PR averaging).
    Previous analyses have either established upper bounds \cite{kaledin2020finite} or established second-moment convergence under the special case that the slow variable is updated with step size $1/k$ \cite{haque2023tightfinitetimebounds}.
    Our result is the first to prove second-moment convergence rates for a general class of step sizes.
    This provides additional insight into the finite-sample behavior of the last iterates and may be of independent interest.

    \item
    Using \Cref{lem:mse}, we establish a finite-time Wasserstein-1 bound in \Cref{thm:clt} for the TSA-PR iterates and their Gaussian limits.
    This is the first non-asymptotic CLT in the context of two-time-scale algorithms.
    As a consequence, we obtain the optimal $K^{-1/2}$ rate of convergence on the expected errors $\mathbb{E}\big[\lVert \bar{x}_K - x^* \rVert\big]$ and $\mathbb{E}\big[\lVert \bar{y}_K - y^*\rVert\big]$ achieved by TSA-PR (\Cref{cor:mae}).
    We prove a lower bound in \Cref{thm:lower_bound} that establishes order-optimality of the $\mathcal{O}(\sqrt{d/K})$ rate.

    \item
    While the asymptotic CLT in \cite{mokkadem2006convergence} characterizes the limiting distribution of TSA-PR, our nonasymptotic analysis provides a finite-time error bound that can be minimized for principled step size selection.
    We show that the time-scale separation $\epsilon_K = \alpha_K/\gamma_K$ that minimizes this bound is asymptotically finite ($\epsilon_K = \Theta(1)$).
    This provides a more fine-grained insight than the infinite-separation ratio that arises from minimizing the asymptotic covariance of the non-averaged iterates.

    \item We support our theoretical findings with experiments that confirm that the stepsize rules suggested by our bounds indeed improve the convergence rate of the TSA with PR averaging.

\end{enumerate}

\section{Related Work}\label{sec:literature}
We compare our contributions with prior works in \Cref{tab:comparison}.
We refer to $(\lVert x_K - x^* \rVert, \lVert y_K - y^* \rVert)$ and $(\lVert \bar{x}_K - x^* \rVert, \lVert \bar{y}_K - y^*\rVert)$ as the errors achieved by last iterate (TSA) and PR-averaged (TSA-PR) algorithms respectively.

\begin{table*}[t]
\centering
\caption{Comparison of analyses for two-time-scale stochastic approximation with and without averaging.
A ``rate of convergence'' guarantees convergence of the scaled second moment $\gamma_K^{-1} \mathbb{E}[ \hat{y}_K \hat{y}_K^T]$ to its limit, which is not ensured by an ``upper bound''.
The matrix $\Sigma_{ss}^\infty$ is used to denote the asymptotic limit of the covariance $\gamma_K^{-1} \mathbb{E} [\hat{y}_K \hat{y}_K^T]$, whose expression is affected by the choice of step sizes.
The $\mathcal{O}(K^{-1/4})$ rate in the last row is achieved by \Cref{thm:clt} for the choice of step size in \eqref{eq:pr_step_size}.
}
\label{tab:comparison}
\setlength{\tabcolsep}{4pt}
\begin{tabular*}{\textwidth}{@{\extracolsep{\fill}}llcll@{}}
\toprule
\textbf{Reference} & \textbf{Output} & \textbf{Slow-Scale Step Size ($\gamma_K \propto K^{-b}$)} & \textbf{Main Result} & \textbf{Remarks} \\ \midrule
\cite[Theorem 1]{kaledin2020finite} & Last-Iterate & $b \in (1/2, 1]$ & $\mathrm{Tr} \mathbb{E}[ \hat{y}_K \hat{y}_K^T ] = \mathcal{O}(\gamma_K)$ & Upper bound \\
\cite[Theorem 4.1]{haque2023tightfinitetimebounds} & Last-Iterate & $b = 1$ & $\lVert \mathbb{E} [\hat{y}_K \hat{y}_K^T]  - \gamma_K \Sigma_{ss}^\infty \rVert = \mathcal{O}(\sqrt{\gamma_K} \alpha_K + \alpha_K^{-1} \gamma_K^2)$ & Rate of convergence \\
\textbf{This Work (\Cref{lem:mse})} & Last-Iterate & $b \in (1/2, 1)$ & $\lVert \mathbb{E} [\hat{y}_K \hat{y}_K^T] - \gamma_K \Sigma_{ss}^\infty \rVert = \mathcal{O}(K^{-1} +  \alpha_K^{-1} \gamma_K^2 )$ & Rate of convergence \\ \midrule
\cite[Theorem 2]{mokkadem2006convergence} & Averaged & $b \in (1/2, 1)$ & $\sqrt{K} \Delta (\bar{y}_K - y^*) \overset{d.}{\to} \bar{\pi}_y$ & Asymptotic CLT \\
\textbf{This Work (\Cref{thm:clt})} & Averaged & $b \in (1/2, 1)$ &
$\dwass (\sqrt{K}\Delta (\bar{y}_K - y^*), \bar{\pi}_y) = \mathcal{O}(K^{-1/4})$ & Finite-sample CLT
\\ \bottomrule
\end{tabular*}
\end{table*}

\subsection{Polyak-Ruppert Averaging and Two-Time-Scale Algorithms}\label{sec:PR_TTSA}
Understanding the optimal rates that can be achieved by a class of algorithms is important for algorithm design.
The study of optimal rates in stochastic approximation has been a central focus since \cite{RobbinsMonro}.
In the context of single-time-scale algorithms, \cite{polyakJuditsky} showed that the asymptotically optimal rate is achieved by the PR averaging scheme proposed by \cite{ruppert,polyak}.
The asymptotic optimality of PR averaging was extended to two-time-scale algorithms by \cite{mokkadem2006convergence}.

All the above results are asymptotic.
The widespread application of stochastic approximation algorithms has spurred interest in understanding finite-time bounds.
For single-time-scale algorithms, finite-time bounds on averaged iterates are known when the step size is constant \cite{mou2020linearstochasticapproximationfinegrained,durmus2024finite}.
These bounds involve correction terms because constant step sizes are not asymptotically optimal.
Finite-time bounds for SA-PR (averaged iterates with decaying step sizes) were established in \cite{srikant19,chen2022finite,srikant2024CLT}.
Less is known for the more general case of two-time-scale algorithms.
Existing finite-time bounds \cite{dalal2018finite,kaledin2020finite,haque2023tightfinitetimebounds} are known only for TSA without the averaging.
Here, we establish the first finite-time bound for the averaged iterates (TSA-PR) generated by two-time-scale algorithms with decreasing step sizes.
The rates achieved by TSA-PR are shown to be significant improvements over those achieved by TSA, akin to the single-time-scale case.

\subsection{Limit Theorems and Quantitative Bounds}
Central limit theorems have been established for both SA and TSA algorithms \cite{polyakJuditsky,mokkadem2006convergence,hu2024central}.
But these results are asymptotic, and therefore cannot be applied to rigorously test the significance of repeated trials that halt after a finite number of iterations.
Non-asymptotic CLTs were established in \cite{anastasiou2019normal,samsonov2024gaussian,srikant2024CLT}, which capture the normality behavior of single-time-scale algorithms.
In this paper (\Cref{thm:clt}), we establish a finite-time bound on the Wasserstein-1 metric for TSA-PR using the martingale CLT in \cite{srikant2024CLT}.
The reason for considering the Wasserstein-1 distance as in \cite{srikant2024CLT} is that weaker notions of distance considered in \cite{anastasiou2019normal,samsonov2024gaussian} are not strong enough to deduce explicit bounds such as the expected error.
In \Cref{cor:mae}, we show that convergence in the Wasserstein-1 distance can be used to establish both a lower and upper bound on the expected error, capturing its correct magnitude up to exact constants.

\section{Model and Preliminaries}\label{sec:preliminaries}
In this paper, we consider two-time-scale stochastic approximation algorithms generated by \eqref{eq:ttsa}.
\revised{The fast-time-scale iterate satisfies $x_k \in \mathbb{R}^{d_f}$ and the slow-time-scale iterate $y_k \in \mathbb{R}^{d_s}$, so that $z_k \coloneqq (x_k, y_k) \in \mathbb{R}^{d}$ with $d = d_f + d_s$.
Accordingly, $A_{ff} \in \mathbb{R}^{d_f \times d_f}$, $A_{fs} \in \mathbb{R}^{d_f \times d_s}$, $A_{sf} \in \mathbb{R}^{d_s \times d_f}$, $A_{ss} \in \mathbb{R}^{d_s \times d_s}$, the deterministic right-hand sides $\vartheta_f \in \mathbb{R}^{d_f}$, $\vartheta_s \in \mathbb{R}^{d_s}$, and the centered noise $W_k \in \mathbb{R}^{d_f}$, $V_k \in \mathbb{R}^{d_s}$.}
While we do not consider the so-called ODE approach here to analyze the system, the assumptions on the system matrices in \eqref{eq:ttsa} are easy to explain by relating the above to a singularly perturbed differential equation; see \cite{borkar2008stochastic}:
\begin{equation*}
\begin{aligned}
    \dot{x}(t) &= - (A_{ff} x(t) + A_{fs} y(t) - \vartheta_f) ,
    \\
    \dot{y}(t) &= - \frac{\gamma}{\alpha} \left(A_{sf} x(t) + A_{ss} y(t) - \vartheta_s\right).
\end{aligned}
\end{equation*}
In the limit $\gamma/\alpha \to 0$, $x(t)$ evolves much faster than $y(t)$.
When the system governing $\dot{x}(t)$ is stable, the slow-time-scale system governed by $\dot{y}(t) = -(A_{sf} x_\infty (y(t)) + A_{ss} y(t)) + \vartheta_s$ is analyzed assuming a stationary solution $x_\infty (y) = A_{ff}^{-1} (- A_{fs} y + \vartheta_f)$:
\begin{equation*}
    \dot{y}(t)
    = -(A_{ss} - A_{sf} A_{ff}^{-1} A_{fs}) y(t) + \vartheta_s - A_{sf} A_{ff}^{-1} \vartheta_f.\vspace{-0.2cm}
\end{equation*}
\subsection{Technical Assumptions}
To ensure that both $x(t)$ and $y(t)$ converge to their respective limits, it is therefore assumed that $-A_{ff}$ and $-\Delta = -(A_{ss} - A_{sf} A_{ff}^{-1} A_{fs})$ are both Hurwitz stable, i.e., the eigenvalues lie in the left-half of the complex plane.
The rates at which the discretized system in \eqref{eq:ttsa} approach their limits are studied under the same setting, which we now state formally.
\begin{assumption}[System Parameters]\label{assumption:structure}\label{assumption:first}
    The matrix $A_{ff}$ and its Schur complement $\Delta = A_{ss} - A_{sf} A_{ff}^{-1} A_{fs}$ are real and satisfy for some positive definite matrices $P_{ff}$ and $P_\Delta$
    \begin{equation*}
        A_{ff}^T P_{ff} + P_{ff} A_{ff} = I, \quad \Delta^T P_\Delta + P_\Delta \Delta = I .
    \end{equation*}
\end{assumption}
\revised{The Hurwitz stability of $-A_{ff}$ ensures stability of the fast subsystem when the slow variable is held constant, while the Hurwitz stability of $-\Delta$ ensures stability of the slow subsystem when the fast variable tracks its equilibrium.
While we formulate our finite-time analysis for linear algorithms, this assumption characterizes local stability of non-linear TSA \cite{mokkadem2006convergence}.
In the non-linear setting, the block matrix $A$ corresponds to the Jacobian evaluated at the root.

\Cref{assumption:structure} holds in standard reinforcement learning applications.
For linear policy evaluation algorithms such as TDC, the fast variable tracks a gradient correction where $A_{ff}$ acts as a positive definite feature covariance matrix, while $\Delta$ forms an effective TD operator \cite{sutton2009fast}.}
For a positive definite matrix $P$ and any matrix $A$, we define the weighted norm
\begin{equation*}
    \lVert A \rVert^2_P = \sup_{x \neq 0}\frac{\langle A x, Ax \rangle_P}{\langle x, x \rangle_P}
    = \sup_{x \neq 0}\frac{x^T A^T P A x}{x^T P x} .
\end{equation*}
Under \Cref{assumption:structure}, it follows that
\begin{equation}\label{eq:contraction}
\begin{split}
    \lVert I - 2 \alpha A_{ff} \rVert_{P_{ff}} &\leq \left(1 - \frac{\mu_{ff}}{2} \alpha \right) ,
    \\
    \lVert I - 2 \gamma \Delta \rVert_{P_\Delta} &\leq
    \left(1 - \frac{\mu_\Delta}{2} \gamma \right) ,
\end{split}
\end{equation}
for some positive constants $\mu_{ff}, \mu_\Delta$ and sufficiently small $\alpha, \gamma > 0$ \cite{kaledin2020finite}.
To study the convergence of \eqref{eq:ttsa}, we consider the following assumption of the stochastic model.

\begin{assumption}[Noise]\label{assumption:noise}
    Let $\{N_k\} \coloneqq \{(W_k, V_k )\}_{k=1}^\infty$ be a martingale difference sequence with respect to $\mathcal{H}_{k} = \{x_1, y_1, W_1, V_1, \cdots, W_{k-1}, V_{k-1}\}$.
    We assume \postrevision{$\sup_{k \geq 1} \mathbb{E}[\lVert N_k \rVert^{2 + \beta}] < \infty$} for some $\beta \in (1/2, 1)$ and that the covariance conditioned on $\mathcal{H}_k$ is
    \begin{align*}
        \mathbb{E}[N_k N_k^T | \mathcal{H}_{k}] = \Gamma = \begin{pmatrix}
            \Gamma_{ff} & \Gamma_{fs} \\ \Gamma_{sf} & \Gamma_{ss}
        \end{pmatrix}
    \end{align*}
    for some positive definite $\Gamma \succ 0$.
\end{assumption}

\revised{While reinforcement learning applications often feature Markovian noise with state-dependent covariances, \Cref{assumption:noise} assumes martingale difference noise with fixed conditional covariance.
When dealing with Markovian noise, it is now standard \cite{srikant2024CLT} to use a Poisson equation to decompose the noise into a martingale difference component and a remainder.
The techniques developed in this paper directly apply to handling this martingale component.
We believe the remaining Markovian terms can be bounded using techniques such as those in \cite{srikant19,srikant2024CLT}, though carrying out this full extension is left for future work.}
Finally, we consider the following class of step sizes.
\begin{assumption}[Step Size]\label{assumption:steps}\label{assumption:last}
    \revised{For any horizon $K \geq 1$,} the step sizes are chosen to be $\alpha_k = \alpha_1 k^{-a}$ and $\gamma_k = \gamma_1 k^{-b}$ with $1/2 < a < b < 1$ and $\gamma_k < \alpha_k$ for all $k \geq 1$.
    Moreover, we assume that for some problem-dependent constants $M_f', M_{fs}', M_s'$, and for all times $k \geq 1$,
    \begin{equation}\label{eq:gap_requirement}
        \frac{\gamma_k}{\alpha_k} \leq \frac{1}{4} \min\left\{\frac{\mu_{ff}}{M_f'}, \frac{\mu_{ff}}{M_{fs}'}, \frac{\mu_\Delta}{M_s'} \right\}.
    \end{equation}
\end{assumption}
The constants $M_f', M_{fs}', M_s'$ are defined in \Cref{app:tsa_mse}.
By the choice of diminishing step sizes $\alpha_k, \gamma_k \to 0$, there exists a time $k_0$ such that \eqref{eq:contraction} holds for all $k > k_0$.
\revised{For every horizon $K$, the exponents must be chosen such that $b > a$,} where the asymptotic time-scale gap is
\begin{equation}\label{eq:time_gap}
    \lim_{K \to \infty} \frac{\alpha_K}{\gamma_K} = \epsilon > 1.
\end{equation}
Note that here we allow $a$ and $b$ to depend on the time horizon $K$ (see Eq.\ \eqref{eq:pr_step_size} below for a detailed discussion of stepsize selections). When $\epsilon = \infty$, the initial step sizes $\alpha_1, \gamma_1$ can be chosen arbitrarily and the analysis in this paper holds for all $k \geq k_0$, where we assume $k_0 = 1$ for simplicity of exposition.
If the asymptotic time-scale gap is finite, then we require that $\alpha_1, \gamma_1$ are chosen such that \eqref{eq:gap_requirement} is satisfied.

\noindent\textbf{Time-Scale Gap $\epsilon$.} The analysis of \eqref{eq:ttsa} has often focused on the convergence rates of the last iterates $(x_K, y_K)$.
In this context, the asymptotic covariance is minimized by enforcing a large time-scale gap $\epsilon$ in \eqref{eq:time_gap}.
On the other hand, the asymptotic covariance of the PR average is invariant to $\epsilon$, and the objective of minimizing the last iterate's covariance, which drives the choice of $\epsilon \to \infty$, is not the relevant criterion for the PR average.
Our key finding is that while the asymptotic covariance of the PR average is invariant to $\epsilon$, its non-asymptotic behavior is not.
Our analysis will demonstrate that controlling the convergence rate in the Wasserstein-1 distance provides a clear theoretical basis for the step size design that provides more insight beyond the convention $\epsilon \to \infty$ based on last-iterate analysis.
We will show that in fact, a mild time-scale gap optimizes the finite-time convergence rate.

\section{Main Results}\label{sec:results}
In this section, we establish a finite-time bound on the Wasserstein-1 distance between the scaled TSA-PR error and its Gaussian limit.
We highlight the practical significance of this result by proving lower- and upper-bounds on the expected error achieved by TSA-PR.
To this end, we first prove that the second moments of TSA error converge to zero at rates determined by the choice of step sizes.

\revised{In analyzing the system, we apply a standard coordinate shift.
Let $(x^*, y^*)$ be the unique fixed point satisfying \eqref{eq:problem}.
By centering the iterates, the analysis can be conducted assuming $\vartheta_f = 0$ and $\vartheta_s = 0$ without loss of generality, effectively reducing the analysis to the case where the optimal solution is $x^* = 0$, $y^* = 0$.
We adopt this convention and assume $\vartheta_f = \vartheta_s = 0$ for all subsequent derivations and in the appendix. Our analysis in this paper, however, can be extended to any arbitrary values of $(x^*, y^*)$.}


Define $\hat{x}_K = x_K - x_\infty(y_K)$ and $\hat{y}_K = y_K - y^*$.
Let $(\Sigma_{ff}, \Sigma_{fs}, \Sigma_{sf}, \Sigma_{ss})$ solve
\begin{equation}\label{eq:covariances}
\begin{split}
    A_{ff} \Sigma_{ff} + \Sigma_{ff} A_{ff}^T &= \Gamma_{ff} , \\
    A_{ff} \Sigma_{fs} + \Sigma_{ff} A_{sf}^T &= \Gamma_{fs} , \\
    \Delta \Sigma_{ss} + \Sigma_{ss} \Delta^T + A_{sf} \Sigma_{fs} + \Sigma_{sf} A_{sf}^T &= \Gamma_{ss}    ,
\end{split}
\end{equation}
with $\Sigma_{fs} = \Sigma_{sf}^T$.
Under the time-scale separation $\epsilon = \infty$ and \Cref{assumption:steps}, it was shown in \cite{konda2004convergence} that the asymptotic covariances of the scaled errors
\begin{gather*}
    \lim_{K \to \infty} \alpha_K^{-1} \mathbb{E}[\hat{x}_K \hat{x}_K^T],
    \quad
    \lim_{K \to \infty} \gamma_K^{-1} \mathbb{E} [\hat{x}_K \hat{y}_K^T] ,
    \\
    \lim_{K \to \infty} \gamma_K^{-1} \mathbb{E}[\hat{y}_K \hat{y}_K^T]
\end{gather*}
are given by $\Sigma_{ff}, \Sigma_{fs}$, and $\Sigma_{ss}$ in \eqref{eq:covariances}.
More generally, it can be shown that $\mathbb{E} [\hat{x}_K \hat{x}_K^T]$ and $\mathbb{E}[\hat{y}_K \hat{y}_K^T]$ converge to a limit when $\epsilon < \infty$, but the asymptotic covariances deviate from those in \eqref{eq:covariances}.

In establishing convergence of the averaged iterates, we must first characterize the finite-time behavior of the second moments of the underlying TSA recursions.
While the asymptotic covariances are known \cite{konda2004convergence}, finite-time bounds for general step sizes \cite{kaledin2020finite} have only established upper bounds on the mean-squared error.
Moreover, the asymptotic covariances \eqref{eq:covariances} may not be the ones achieved by the last iterates under the general choice of step sizes in \Cref{assumption:steps}.
While \cite{haque2023tightfinitetimebounds} established convergence rates for the covariance matrices, their analysis is restricted to the special case that $\gamma_K$ is chosen to be proportional to $1/K$.
\Cref{lem:mse} bridges this gap by providing an explicit convergence rate for the second moment matrices under a general class of step sizes.
This is important for establishing \Cref{thm:clt}, where the rate of convergence of the averaged iterates is shown to benefit from the choice of step size $\gamma_K \propto K^{-b}$ with $b < 1$.

\postrevision{For symmetric $R$ and $P \succ 0$, define the norm $\lVert R\rVert_{P,\mathrm F}=\lVert P^{1/2}RP^{1/2}\rVert_{\mathrm F}$, where $\| \cdot \|_F$ is the Frobenius norm.}
Define
\begin{equation}\label{eq:def_delta_y}
\delta_y(k) \coloneqq
\postrevision{\lVert \mathbb{E}[ \hat{y}_k \hat{y}_k^T]- \gamma_k \Sigma_{ss} \rVert_{P_\Delta,\mathrm F}}
.
\end{equation}
Following \cite{konda2004convergence}, let $\{L_k\}$ be a sequence satisfying
\begin{equation*}
\begin{split}
    &L_{k+1} (I - \gamma_k (\Delta - A_{sf} L_k))
    \\ = &
    L_k - \alpha_k A_{ff} L_k + \gamma_k A_{ff}^{-1} A_{fs}(\Delta - A_{sf} L_k)
\end{split}
\end{equation*}
with $L_0 = 0$.
From \cite[Lemmas 18--19]{kaledin2020finite}, $L_k$ satisfies $\lVert L_k \rVert = \mathcal{O}(\gamma_k/\alpha_k)$.
Define $\tilde{L}_{k+1} \coloneqq L_{k+1} + A_{ff}^{-1}A_{fs}$.
Using $L_k$, the recursions for $\tilde{x}_k = \hat{x}_k + L_k \hat{y}_k$ and $\hat{y}_k$ can be decoupled as
\begin{equation}\label{eq:fast_slow_updates}
    \begin{split}
    \tilde{x}_{k+1} &= (I - \alpha_k A_{ff}) \tilde{x}_k - \gamma_k \tilde{L}_{k+1} A_{sf} \tilde{x}_k
    \\ &\quad + \alpha_k W_k + \gamma_k \tilde{L}_{k+1} V_k,
    \\
    \hat{y}_{k+1} &= (I - \gamma_k \Delta) \hat{y}_k + \gamma_k A_{sf} L_k \hat{y}_k - \gamma_k A_{sf} \tilde{x}_k + \gamma_k V_k.
    \end{split}
\end{equation}
We establish finite-time bounds on $\delta_y(k)$ and
\begin{equation}\label{eq:def_delta_tilde_x}
    \tilde{\delta}_x (k)
    \coloneqq
    \postrevision{\left\lVert \mathbb{E}[\tilde{x}_k \tilde{x}_k^T] - \alpha_k \Sigma_{ff} \right\rVert_{P_{ff},\mathrm F}}
    ,
\end{equation}
which measure the deviations of $\mathbb{E}[\hat{y}_k \hat{y}_k^T]$ and $\mathbb{E}[\tilde{x}_k \tilde{x}_k^T]$ from $\gamma_k \Sigma_{ss}$ and $\alpha_k \Sigma_{ff}$, respectively.
\begin{lemma}\label{lem:mse}
    Under \Crefrange{assumption:first}{assumption:last} and for some problem-dependent constants $M_f, M_s > 0$, the following holds for every $K \geq 1$,
    \begin{equation}\label{eq:lemma1_bounds}
    \begin{split}
        \tilde{\delta}_x (K+1) &\leq  \prod_{k=1}^K \left(1 - \alpha_k \frac{\mu_{ff}}{4}\right) \tilde{\delta}_x (1) + M_f \gamma_K , \\
        \delta_y (K+1) &\leq \prod_{k=1}^K \left(1 - \gamma_k \frac{\mu_\Delta}{4}\right) \delta_y (1) + M_s \left(\frac{1}{K} + \frac{\gamma_K^2}{\alpha_K} \right).
    \end{split}
    \end{equation}
\end{lemma}
\begin{proof}
    The proof is provided in Appendix \ref{app:tsa_mse}.
\end{proof}
The bounds in \eqref{eq:lemma1_bounds} provide rates of convergence for $\mathbb{E}[\tilde{x}_k \tilde{x}_k^T]$ and $\mathbb{E}[\hat{y}_k \hat{y}_k^T]$ to their asymptotic covariances when $\gamma_K/\alpha_K \to 0$ and $1/K\gamma_K \to 0$.
When $\alpha_K$ and $\gamma_K$ are chosen such that $\gamma_K/\alpha_K$ converges to a non-zero limit, the asymptotic covariance $\lim_{K \to \infty} \postrevision{\gamma_K^{-1}}\mathbb{E} [\hat{y}_K \hat{y}_K^T]$ deviates from $\Sigma_{ss}$ defined in \eqref{eq:covariances}.
Nevertheless, the bound \eqref{eq:lemma1_bounds} implies $\lVert \mathbb{E} [ \hat{y}_K \hat{y}_K^T ]\rVert = \mathcal{O}(\gamma_K)$, which is used to establish convergence of the averaged iterates.

Recall that the Wasserstein-1 distance is defined for two random variables $X$ and $Z$ as
\begin{equation}\label{eq:wasserstein_definition}
    \dwass \left(X, Z\right) \coloneqq
    \sup_{h \in \mathrm{Lip}_1} \mathbb{E}\left[h(X) - h(Z) \right] .
\end{equation}

\postrevision{Following \cite{mokkadem2006convergence}, we assume for simplicity that $A_{ss}$ is invertible so that the asymptotic distribution of TSA-PR can be expressed using the Schur complement $G = A_{ff} - A_{fs} A_{ss}^{-1} A_{sf}$ of $A_{ss}$, as well as the Schur complement $\Delta$ defined in \Cref{sec:preliminaries}.
When $A_{ss}$ is singular, the bound in \Cref{thm:clt} instead applies to the joint scaled error $\sqrt{K}(\bar{x}_K - x^*, \bar{y}_K - y^*)$, whose Gaussian limit has covariance $A^{-1} \Gamma A^{-T}$.
}
\revised{The distinct roles of the Schur complements $G$ and $\Delta$ arise from the structure of the joint system matrix $A = \big[\begin{smallmatrix} A_{ff} & A_{fs} \\ A_{sf} & A_{ss} \end{smallmatrix}\big]$.
While the non-averaged fast iterate $x_k$ is locally governed by $A_{ff}$ due to the time-scale separation, Polyak-Ruppert averaging achieves the optimal asymptotic covariance characterized by $A^{-1}$.
By block matrix inversion, the upper-left and lower-right blocks of $A^{-1}$ are exactly $G^{-1}$ and $\Delta^{-1}$.
Consequently, $G$ and $\Delta$ act as the effective linear operators that define the asymptotic covariances of the averaged fast and slow iterates, respectively.}

Utilizing the error bounds in \Cref{lem:mse} for TSA, we establish a non-asymptotic CLT for $\sqrt{K}(\bar{x}_K - x^*)$ and $\sqrt{K}(\bar{y}_K - y^*)$ to their Gaussian limits by obtaining a bound on the Wasserstein-1 distance.
Let $\bar{\Sigma}_{ff}$ and $\bar{\Sigma}_{ss}$ be solutions to the matrix equations
\begin{equation}\label{eq:pr_covariance}
\begin{split}
    G \bar{\Sigma}_{ff} G^T &=  \lim_{K \to \infty} \frac{1}{K} \mathrm{Cov}\left( \sum_{k=1}^K (W_k - A_{fs} A_{ss}^{-1} V_k) \right),
    \\
    \Delta \bar{\Sigma}_{ss} \Delta^T &= \lim_{K \to \infty} \frac{1}{K} \mathrm{Cov}\left(\sum_{k=1}^K (V_k - A_{sf} A_{ff}^{-1} W_k )\right) .
\end{split}
\end{equation}
\revised{The unbarred matrices $(\Sigma_{ff},\Sigma_{ss})$ in \eqref{eq:covariances} characterize the last iterates, whereas the barred matrices $(\bar{\Sigma}_{ff},\bar{\Sigma}_{ss})$ in \eqref{eq:pr_covariance} characterize TSA-PR; these covariances generally differ because they solve distinct equations.}

Let $\bar{\pi}_x = \mathcal{N}(0, G \bar{\Sigma}_{ff} G^T)$ and $\bar{\pi}_y = \mathcal{N}(0, \Delta \bar{\Sigma}_{ss} \Delta^T)$.
\begin{theorem}\label{thm:clt}
    \revised{Under Assumptions \ref{assumption:first}--\ref{assumption:last}} \postrevision{with $A_{ss}$ invertible}, when the step sizes $a, b$ are chosen in the set
    \begin{equation*}
        \mathcal{E} = \left\{(a, b):
         \frac{1}{2}  < a < b < 2a - \frac{1}{2}
         \right\}
    \end{equation*}
    then $\dwass(\sqrt{K} G(\bar{x}_K - x^*), \bar{\pi}_x)$ and $\dwass(\sqrt{K} \Delta (\bar{y}_K - y^*), \bar{\pi}_y)$ converge to zero at rate
    \begin{equation*}
            \mathcal{O}\left(\frac{1}{\sqrt{K}} \left(\frac{K^{a-1/2}}{a-1/2} + \frac{1}{b-a} + K^{b/2}
                +\frac{K^{2a - b - 1/2}}{2a  - b - 1/2}
            \right)
            \right) .
    \end{equation*}\vspace{0.1cm}
\end{theorem}
\begin{proof}
The proof is provided in Appendix \ref{app:clt}.
\end{proof}
\revised{In the bound above, the $\mathcal{O}(\cdot)$ notation suppresses problem-dependent constants that are independent of the horizon $K$.
Specifically, these constants depend on the system matrices through the stability constants ($\mu_{ff}, \mu_\Delta, P_{ff}, P_\Delta$), the dimension $d$, the initial errors ($\mathbb{E}[\|x_1\|], \mathbb{E}[\|y_1\|]$), the step-size coefficients ($\alpha_1, \gamma_1, c_a, c_b$), and the noise moments.
We omit the explicit algebraic expressions for these terms; tracking these constants exactly through the nested recursions yields lengthy expressions that are not optimized with respect to these system parameters, whereas our primary focus is characterizing the exact finite-time rates with respect to $K$.}

\textbf{Technical Insights of \Cref{thm:clt}.} \Cref{thm:clt} presents a simplified bound that is valid when the step sizes lie in the subset $\mathcal{E}$ of \Cref{assumption:steps}.
The full error bound for all $a, b$ satisfying \Cref{assumption:steps} is provided in Appendix \ref{app:clt}.
The subset of step sizes specified by $\mathcal{E}$ is highlighted in the main body of the paper because they optimize the error bound.

This error bound provides a path to selecting step size exponents for TSA-PR. The step size exponents $a$ and $b$ should be chosen to balance the polynomial decay rates with the singular behavior of the coefficients, i.e., terms such as $1/(a-1/2)$ and $1/(b-a)$.
In particular, the choice
\begin{equation}\label{eq:pr_step_size}
    a = \frac{1}{2} + \frac{c_a}{\log K} \qquad \text{and}
    \qquad
    b = \frac{1}{2} + \frac{c_b}{\log K}
\end{equation}
for constants $0 < c_a < c_b < 2 c_a$ optimizes the error bound.
\revised{We note that \eqref{eq:pr_step_size} is a finite-horizon tuning rule permitted by the horizon-dependent formulation of \Cref{assumption:steps}; an anytime or horizon-free version is not claimed.}
The choice \eqref{eq:pr_step_size} leads to much better bounds than choosing $a, b$ independent of $K$\revised{: fixed exponents close to $1/2$, e.g., $a = 1/2 + \delta$, are admissible but inject a strictly suboptimal $K^\delta$ factor into the error}.
An implication of this step size choice concerns the effective time-scale separation at the terminal step,
\begin{equation*}
    \epsilon = \frac{\alpha_1}{\gamma_1} \lim_{K \to \infty} K^{(c_b - c_a)/\log K} = \frac{\alpha_1}{\gamma_1} e^{c_b - c_a} .
\end{equation*}
\revised{This finite terminal ratio $\epsilon$ is a distinct finite-horizon effect.
It is notable when contrasted with fixed-exponent asymptotic regimes, which enforce an infinite time-scale gap $\epsilon = \infty$ to achieve the optimal asymptotic limit of TSA without averaging.}
Our analysis indicates that for PR averaging, such a large separation may not be optimal.
The averaging mechanism benefits from a much finer separation between the time scales, which leads to faster convergence to the optimal asymptotic covariance.

To our knowledge, \cite{mokkadem2006convergence} is the only work where the authors analyzed the performance of PR averaging in the context of two-time-scale algorithms (the authors proved their asymptotic behavior).
Our analysis builds on the proof technique, where the PR averages are expressed as
\begin{equation}\label{eq:pr_expression_fast}
\begin{split}
        & G \bar{x}_K  - \frac{1}{K} \sum_{k=1}^K (W_k - A_{fs} A_{ss}^{-1} V_k)
        \\ = &  \frac{1}{K} \sum_{k=1}^K \left(\alpha_k^{-1} (x_{k} - x_{k+1}) -  \gamma_k^{-1} A_{fs} A_{ss}^{-1} (y_k - y_{k+1}) \right),
\end{split}
\end{equation}
and
\begin{equation}\label{eq:pr_expression_slow}
\begin{split}
        &\Delta \bar{y}_K - \frac{1}{K} \sum_{k=1}^K (V_k - A_{sf} A_{ff}^{-1} W_k)
        \\ =&   \frac{1}{K}\sum_{k=1}^K \left(\revised{- A_{sf} A_{ff}^{-1} \alpha_k^{-1} (x_{k} - x_{k+1})}
        +  \gamma_k^{-1} (y_{k} - y_{k+1}) \right)  .
\end{split}
\end{equation}
These expressions decompose the TSA-PR error into a sum of a standardized martingale and remainder terms.
The convergence rate of the Wasserstein-1 distance is determined by the decay rate of these remainder terms.
A preliminary analysis based on the bounds from \Cref{lem:mse} via Jensen's inequality leads to the convergence rate $\mathcal{O}(K^{-1/4})$ as $(a, b) \to (1/2, 1/2)$.
However, this argument is insufficient for a rigorous finite-time analysis.
Such an approach overlooks the fact that the error can exhibit singular behavior at some points on the set $[1/2, 1]^2$.
Singularity points arise in the weighted sums in \eqref{eq:pr_expression_fast} and \eqref{eq:pr_expression_slow}, which can amplify the effect of transient rates in the last-iterate analysis.
For instance, we show that the bounds depend critically on quantities such as $(b-a)^{-1}$ and $(a-1/2)^{-1}$.
A direct analysis that ignores these coefficients fails to reveal how large the error can grow on such singularity points.
A fine-grained analysis is therefore essential to characterize PR average's rate of convergence.
Through a careful analysis, we find that the total error bound undergoes several phase transitions throughout the choice of step sizes $a, b \in (1/2, 1)$.
\Cref{thm:clt} is a result of this detailed analysis, formalizing the structure of the most critical terms that govern the optimization of the error bound.

Recall the Kantorovich--Rubinstein dual form of the Wasserstein-1 distance in \eqref{eq:wasserstein_definition}.
Since this is a supremum over all $1$-Lipschitz test functions, evaluating it at the maximum over the two functions $h(x) \in \{-\lVert x \rVert, \lVert x \rVert\}$, we have the following lower and upper bounds on the expected error
\begin{equation*}
\begin{split}
    &\left\lvert \mathbb{E}\!\left[\lVert G(\bar{x}_K - x^*)\rVert\right]
    - \tfrac{1}{\sqrt{K}}\,\mathbb{E}\!\left[\lVert Z_1 \rVert\right]\right\rvert
    \\
    &\leq \tfrac{1}{\sqrt{K}}\, \dwass\!\left(\sqrt{K}\,G(\bar{x}_K - x^*),\,\bar{\pi}_x\right),
\end{split}
\end{equation*}
with the lower bound obtained from the test function $-\lVert x\rVert$ and the analogous statement holding for the slow component.
\begin{corollary}\label{cor:mae}
    Let $Z_1 \sim \bar{\pi}_x$ and $Z_2 \sim \bar{\pi}_y$, where $\bar{\pi}_x$ and $\bar{\pi}_y$ are defined in \Cref{thm:clt}.
    Under \Crefrange{assumption:first}{assumption:last},
    \begin{equation*}
        \begin{split}
            \left\lvert \mathbb{E}[\lVert G(\bar{x}_K - x^*) \rVert] - \frac{1}{\sqrt{K}} \mathbb{E}[\lVert Z_1 \rVert]
            \right\rvert = o(K^{-1/2})
            \\
            \left\lvert \mathbb{E}[\lVert \Delta (\bar{y}_K - y^*) \rVert] - \frac{1}{\sqrt{K}} \mathbb{E}[\lVert Z_2 \rVert] \right\rvert = o(K^{-1/2})
        \end{split}
        .
    \end{equation*}\vspace{0.1cm}
\end{corollary}
\revised{In particular, under the step sizes in \eqref{eq:pr_step_size}, the Wasserstein-1 distance in \Cref{thm:clt} decays at $\mathcal{O}(K^{-1/4} )$.
Consequently, the expected error decays with a leading rate of $\Theta(K^{-1/2})$, while the remainder (correction term) bounded in \Cref{cor:mae} decays at $\mathcal{O}(K^{-3/4})$.}
The above bounds are tight in the sense that they provide upper and lower bounds on $\mathbb{E}[\lVert G(\bar{x}_K - x^* )\rVert]$ and similarly for $\bar{y}_K - y^*$.
Also note that the exact constant for the expected error is correctly captured; replacing $\mathbb{E}[\lVert Z_1 \rVert]$ with $\sqrt{\mathrm{Tr}\mathrm{Cov}(Z_1)}$ would result in a $\mathcal{O}(K^{-1/2})$ error instead of the $o(K^{-1/2})$ above.

\subsection{Optimality of Polyak-Ruppert Averaging}\label{sec:optimality}
For single-time-scale algorithms, it is known that PR averaging achieves the optimal (smallest) asymptotic covariance.
It is insightful to note that the asymptotic covariances for TSA-PR \eqref{eq:pr_covariance} admit a clean interpretation.
Stochastic approximation can be used without knowledge of the system matrix $A$, where the variables are updated using noisy observations as in \eqref{eq:ttsa}.
If the true operator $A$ and the driving noise sequence $\{(W_k, V_k)\}_{k=1}^K $ are observed directly, the solution $\hat{z}_K^*$ to the underlying linear system $A \hat{z}_K^* = K^{-1} \sum_{k=1}^K (W_k, V_k)^T$ is associated with the covariance matrix $\Sigma^* = A^{-1} \Gamma A^{-T}$.
This covariance matrix is identical to the asymptotic covariance $\bar{\Sigma}$ achieved by the PR average, demonstrating that the PR average successfully learns the underlying system from noisy samples.

A comparison of the asymptotic covariances $\bar{\Sigma}_{ff}, \bar{\Sigma}_{ss}$ with a class of algorithms shows that TSA-PR is asymptotically optimal \cite{mokkadem2006convergence} in minimizing the mean square errors.
The covariances also match the lower bound as discussed in \cite{tsypkin1974attainable,polyakJuditsky}, or the Cram\'{e}r-Rao lower bound (when TSA-PR is unbiased) as discussed in \Cref{sec:LowerBound}.
This is in contrast to the asymptotic covariances of TSA without averaging, where the asymptotic covariances are at least as large as the matrices defined in \eqref{eq:covariances}.
Next, we discuss improvements and optimality with respect to the expected error $\mathbb{E}[\lVert \cdot \rVert]$.

\Cref{cor:mae} shows that TSA-PR achieves the error bounds
\begin{align*}
    \mathbb{E}[\lVert \bar{x}_K - x^* \rVert]  &\leq \frac{1}{\sqrt{K}} \mathbb{E}[ \lVert G^{-1} Z_1 \rVert] + o(K^{-1/2}),
    \\
    \mathbb{E}[\lVert \bar{y}_K - y^* \rVert ]
    &\leq \frac{1}{\sqrt{K} }\mathbb{E}[\lVert \Delta^{-1} Z_2 \rVert] + o(K^{-1/2}),
\end{align*}
which are much smaller than the errors
\begin{align*}
    \mathbb{E}[\lVert x_K - x^* \rVert] &= \mathcal{O}\left(\sqrt{\alpha_K \mathrm{Tr} \Sigma_{ff}}  \right),
    \\
    \mathbb{E}[\lVert y_K - y^* \rVert] &= \mathcal{O}\left(\sqrt{\gamma_K \mathrm{Tr} \Sigma_{ss}} \right)
\end{align*}
in \Cref{lem:mse} achieved by TSA without averaging, since $K^{-1/2} \ll \sqrt{\gamma_K} \ll \sqrt{\alpha_K}$.
Therefore, averaging reduces errors significantly.

We show that the $K^{-1/2}$ rate achieved by TSA-PR is optimal.
The following lower bound applies to any algorithm that estimates the solution to $A z = \mathbf{\vartheta}$ from $K$ noisy observations $\mathbf{\vartheta} + N_k$ with centered noise $N_k$.
For two-time-scale algorithms, $z$ is to be viewed as the solution $(x, y)$ to \eqref{eq:problem} and $\mathbf{\vartheta} = (b_f, b_s)$.
We consider the general case where the right-hand side $\mathbf{\vartheta}$ is not necessarily zero.
\begin{theorem}\label{thm:lower_bound}
    Consider any estimator $\hat{z}_K$ of $z \in \mathbb{R}^d$ solving $A z = \mathbf{\vartheta}$, with access to $K$ i.i.d. samples $\mathbf{\vartheta} + N_k$.
    \postrevision{Suppose $N_k \overset{\mathrm{i.i.d.}}{\sim} \mathcal{N}(0, \Gamma)$ with} $\Sigma^* = A^{-1} \Gamma A^{-T}$.
    When $d > 48 \log 2$, there exists a \postrevision{finite set $\mathcal{Z}_K \subset \mathbb{R}^d$ such that}
    \begin{equation}\label{eq:high_dimension_lb}
        R_K^* \coloneqq \min_{\hat{z}_K} \max_{z \in \mathcal{Z}_K} \mathbb{E}[\lVert \hat{z}_K - z \rVert]
        \geq
        \frac{1}{3} \sqrt{\frac{d}{128 K} \frac{\lVert \Sigma^* \rVert}{\kappa(\Sigma^*)}}
        .
    \end{equation}
\end{theorem}
\begin{proof}
    The proof is provided in Appendix \ref{sec:LowerBound}, where we also obtain a result when $d \leq 48 \log 2$.
\end{proof}
Let $\bar{R}_K = \postrevision{\max_{z^* \in \mathcal{Z}_K}}\mathbb{E}[\lVert \bar{z}_K - z^* \rVert]$ be the error achieved by TSA-PR in \Cref{cor:mae}.
\postrevision{Absorbing} the transient term $o(K^{-1/2})$ in \Cref{cor:mae} \postrevision{into a constant $C_2$} and applying Jensen's inequality to obtain an upper bound, and comparing with the lower bound $R_K^*$ in \Cref{thm:lower_bound}, we have for constants $C_1, C_2 > 0$\postrevision{, independent of $K$,} the inequalities
\begin{equation*}
    C_1 \sqrt{\frac{d}{K} \frac{\lVert \Sigma^* \rVert}{\kappa(\Sigma^*)}} \leq
    R_K^*
    \leq
    \bar{R}_K
    \leq
    \revised{C_2 \sqrt{\frac{\mathrm{Tr} \Sigma^*}{K} }}
    .
\end{equation*}
\revised{Noting that $\mathrm{Tr} \Sigma^* \leq d \lVert \Sigma^* \rVert$, the upper and lower bounds match in their $\sqrt{d/K}$ scaling.
This demonstrates that TSA-PR achieves order-optimality in $K$ and $d$ relative to an i.i.d. oracle lower bound, up to constants and conditioning factors, despite not having access to the model $A$ or direct observations of the noise $\{N_k\}_{k=1}^K$.}

\section{Experiments}\label{sec:experiments}
In this section, we complement our theoretical results with simulations that illustrate how the error bounds accurately describe the behavior of TSA and TSA-PR in practice.

\begin{figure}[thbp]
    \centering
    \includegraphics[width=\linewidth]{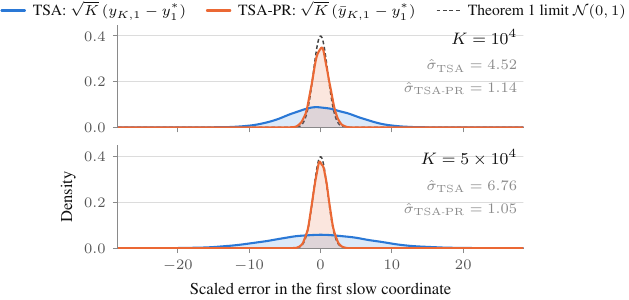}
    \caption{
        Gaussian behavior of the optimality gaps $\sqrt{K}(y_{K, 1} - y_1^*)$ and $\sqrt{K}(\bar{y}_{K, 1} - y_1^*)$.
        The variance of TSA increases with $K$ as suggested by \Cref{lem:mse}, whereas that of TSA-PR converges to 1.
        }
    \label{fig:clt}
\end{figure}

\begin{figure}[htbp]
    \centering
    \includegraphics[width=\linewidth]{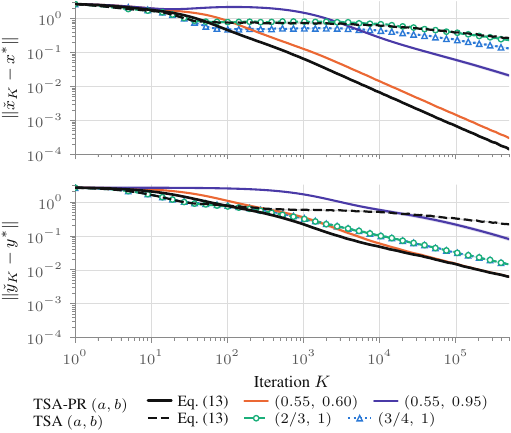}
    \caption{
    Errors achieved by TSA and TSA-PR for policy evaluation.
    Top: fast-time-scale. Bottom: slow-time-scale.
    Step size decay \eqref{eq:pr_step_size} prescribed by Theorem~\ref{thm:clt} achieves the smallest error.
    The legend indicates the value of $(a, b)$ chosen: values are chosen according to either \eqref{eq:pr_step_size} or as specified in the legend.
    }
    \label{fig:TDC}
\end{figure}

\revised{
\Cref{fig:clt,fig:TDC} include TSA-PR using \eqref{eq:pr_step_size}, where $a, b$ are fixed using the terminal horizon. 
The noise $\{(W_k, V_k)\}$ is generated by first sampling i.i.d. variables $\{\varepsilon_k\}$ whose coordinates are i.i.d. $\mathrm{Exp}(1) - 1$ and computing $(W_k, V_k)$ as $L \varepsilon_k$, where $\Gamma = L L^T$. 
We observed that the performance is not sensitive to the choice of $c_a$ and $c_b$ when $c_b$ is set as $c_b = 1.5 c_a$, so we fixed $c_a = 0.1$ and $c_b = 0.15$.
}

\textbf{Synthetic System:}
Our first set of experiments uses randomly generated system parameters $A_{ff},A_{fs},A_{sf},A_{ss}\in\mathbb{R}^{5\times 5}$ subject to the constraints in \Cref{assumption:structure}.
According to \Cref{lem:mse,thm:clt}, the optimality gap $\sqrt{K} (x_K-x^*, y_K-y^*)$ of TSA exhibits increasing variance as $K$ grows, whereas the gap $\sqrt{K} (\bar{x}_K-x^*, \bar{y}_K-y^*)$ of TSA-PR converges to a Gaussian distribution.
\Cref{fig:clt} illustrates the empirical distributions at two different values of $K$ using smoothed kernel density plots.
For simplicity of illustration, we only plot the first coordinates of the slow variables.
The figure reveals that $\sqrt{K} (\bar{y}_K-y^*)$ indeed converges to a Gaussian distribution with standard deviation $1$, which is the asymptotic covariance predicted by \Cref{thm:clt}. The distribution of $\sqrt{K} (y_K-y^*)$ exhibits increasing standard deviation over time as described in \Cref{lem:mse}.

\noindent\textbf{Gradient-Based Policy Evaluation.}
Temporal difference learning with gradient correction (TDC) is a gradient-based policy evaluation algorithm in RL which works stably with function approximation and off-policy sampling \cite{sutton2009fast}.
The algorithm seeks to solve the system of equations
\begin{align*}
    \mathbb{E}[\phi\phi^T]\, x
    + \mathbb{E}[\phi(\phi - \varrho\phi')^T]\, y &= \mathbb{E}[r\phi] , \\
    \varrho\, \mathbb{E}[\phi'\phi^T]\, x
    + \mathbb{E}[\phi(\phi - \varrho\phi')^T]\, y &= \mathbb{E}[r\phi] ,
\end{align*}
where $\phi = \phi(s)$ is the feature vector associated with state $s$, $\phi' = \phi(s')$ is the next state's feature vector, $r = r(s, a)$ is the reward, and $\varrho$ is the discount factor.
Details on how TDC can be expressed in the form \eqref{eq:ttsa} can be found in \cite{xu2019two,zeng2024fast}.

\revised{
In \Cref{fig:TDC}, we compare errors $\lVert \check{x}_K - x^*\rVert,\lVert \check{y}_K - y^*\rVert$ achieved by TSA ($\check{x}_K = x_K, \check{y}_K = y_K$) and TSA-PR ($\check{x}_K = \bar{x}_K, \check{y}_K = \bar{y}_K$) under various step size schedules, showing how different step sizes lead to different convergence rates.
We generate an off-policy policy-evaluation problem with 20 states, 3 actions, and 10-dimensional linear features.
Transition probabilities are independently drawn from a symmetric Dirichlet distribution, rewards and features from standard Gaussian distributions, and the features are whitened under the behavior-policy stationary distribution.
The target policy is a softmax of Gaussian logits, the behavior policy is a uniform distribution over actions, and the discount factor is 0.9.
We use the resulting expected TDC dynamics and replace the sample-dependent update noise by independent additive noise with the same fixed-point covariance.
}

\revised{We include both TSA (last iterate) and TSA-PR with various time-scale gaps.
For TSA, we use step sizes analyzed in \cite{dalal2018finite,kaledin2020finite,haque2023tightfinitetimebounds}.
We also include the errors when \eqref{eq:pr_step_size} is used without averaging, for reference. 
For TSA-PR, we empirically validate performance under various time-scale gaps.
\Cref{fig:TDC} shows that both slow- and fast-time-scale errors enjoy smaller errors under the theoretically prescribed step sizes, with the best TSA error being $2.31 \times$ larger than TSA-PR with \eqref{eq:pr_step_size}.}

\begin{figure}[htbp]
    \centering
    \includegraphics[width=\linewidth]{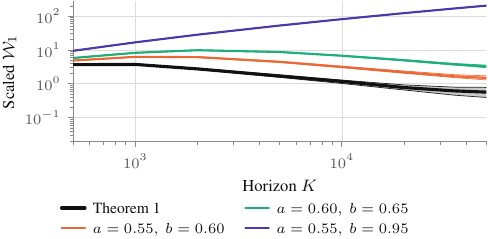}
    \caption{
    Wasserstein-1 distance between $\sqrt{K}(\bar{y}_{K, 1} - y_1^*)$ and its Gaussian limit $\mathcal{N}(0,1)$ normalized by the $K^{-1/4}$ rate in \Cref{thm:clt} corresponding to the step size in \eqref{eq:pr_step_size}.
    The $\mathcal{W}_1$ distance converges fastest at every horizon when using the prescribed step size.
    }
    \label{fig:W1}
\end{figure}

\revised{
We next measure the tightness of \Cref{thm:clt} with respect to the rate $K$, following the setup used for \Cref{fig:clt}.
For each horizon $K$ we estimate $\mathcal{W}_1 \bigl(\sqrt{K}(\bar{y}_{K,1} - y_1^*), \mathcal{N}(0,1)\bigr)$ from $2\times10^4$ independent runs and plot in \Cref{fig:W1} the distance normalized by the predicted rate, as $K^{1/4}\mathcal{W}_1$.
We report the first slow coordinate because in one dimension $\mathcal{W}_1$ is computed exactly from the empirical quantile function.
As shown, the step size prescribed in \eqref{eq:pr_step_size} achieves $K^{-1/4}$ rate, while step sizes with infinite time-scale gaps converge more slowly.
The figure demonstrates that the rate established in the theorem is tight, and highlights the importance of using horizon-dependent step size in \eqref{eq:pr_step_size}.
}



\section{Proof Sketches for \texorpdfstring{\Cref{lem:mse,thm:clt}}{Lemma~\ref{lem:mse} and Theorem~\ref{thm:clt}}}\label{sec:sketch}
The analysis of two-time-scale algorithms is difficult because of the coupling between the fast- and slow-time-scale iterates.
Methods for establishing finite-time MSE bounds on TSA are often based on a decoupling technique introduced by \cite{konda2004convergence}.
The complexity involved in the MSE analysis is further accentuated when analyzing the averaged iterates, because it involves the inner product between the averages which can be intractable.
We observe that the Wasserstein-1 distance between errors achieved by TSA-PR and the limiting Gaussian variables can be made tractable by decomposing the TSA-PR errors as standardized martingale sums and weighted averages of TSA following \cite{mokkadem2006convergence}.

\textbf{Proof of \Cref{lem:mse}:}
Both our \Cref{lem:mse} and \cite[Theorem 4.1]{haque2023tightfinitetimebounds} advance beyond traditional finite-time bounds on the second moments, such as $\mathbb{E}[\lVert \tilde{x}_k \rVert^2]$, by establishing the explicit convergence of the covariance matrices.
However, the underlying methodologies and scope are distinct.
The approach in \cite{haque2023tightfinitetimebounds} can be described as a residual analysis: it examines the recursion for the covariance $\mathbb{E}[\tilde{x}_k \tilde{x}_k^T]$, groups all non-leading terms into a remainder, and then shows this remainder is asymptotically negligible.
In contrast, our method follows from a simplified Lyapunov-style argument, where we show that $\tilde{\delta}_x (k)$ and $\delta_y(k)$ in equations \eqref{eq:def_delta_tilde_x} and \eqref{eq:def_delta_y} undergo contractions:
\begin{equation*}
    \begin{split}
        \tilde{\delta}_x (k+1) &\leq \left(1 - \alpha_k \frac{\mu_{ff}}{4}\right) \tilde{\delta}_x (k) + \mathcal{O}\left(\alpha_k \gamma_k \right)
            ,
        \\
        \delta_y(k+1) &\leq \left(1 - \gamma_k \frac{\mu_\Delta}{4}\right) \delta_y(k) + \mathcal{O}\left(\gamma_k \left(\frac{1}{k} + \frac{\gamma_k^2}{\alpha_k} \right) \right).
    \end{split}
\end{equation*}
This methodological distinction allows our analysis to cover general step sizes, whereas the results in \cite{haque2023tightfinitetimebounds} are specific to the slow-time-scale step size $\gamma_k = \gamma_1 \cdot k^{-1}$.
This generality is crucial for our primary goal of analyzing TSA-PR, which requires carefully tracking the precise convergence rates of all terms across different step size regimes.
As we will show, this detailed analysis enables us to identify certain singularities that arise in the error bounds of PR averaging.

\textbf{Proof of \Cref{thm:clt}:}
Without loss of generality, let $x^* = 0$ and $y^* = 0$.
Using the expression for the PR averages in \eqref{eq:pr_expression_fast} and \eqref{eq:pr_expression_slow} scaled by $\sqrt{K}$, we see that the first terms are standardized sums of martingales.
To obtain their rates of convergence, we use the following quantitative bound, obtained from using \cite[Theorem 1]{srikant2024CLT} with $\beta > 1/2$ in \Cref{assumption:noise}, on the Wasserstein-1 distance between a standardized sum of martingales and its limiting Gaussian variable.
\begin{lemma}\label{lem:CLT}
    Let $\{N_k\} \subset \mathbb{R}^d$ be a martingale difference sequence satisfying \Cref{assumption:noise}.
    For a standard Gaussian vector $Z$, it holds that
    \begin{equation*}
        \dwass \left(K^{-1/2} \sum_{k=1}^K N_k , \Gamma^{1/2} Z \right) \leq \mathcal{O}\left(  \frac{1}{K^{1/4}} \right)
        ,
    \end{equation*}
    where the $\mathcal{O}(\cdot)$ notation hides problem-dependent constants such as the dimension and noise moments.
\end{lemma}

Next, we relate the telescoped terms in equations \eqref{eq:pr_expression_fast} and \eqref{eq:pr_expression_slow} to their respective errors in expectation as
\begin{equation*}
    \begin{split}
        &\mathbb{E}\left[\left\lVert \sum_{k=1}^K \alpha_k^{-1} (x_k - x_{k+1}) \right\rVert \right]
            \\ \leq &
        \frac{\mathbb{E} \left[\lVert x_1 \rVert \right]}{\alpha_1} + \frac{\mathbb{E}\left[\lVert x_{K+1} \rVert\right]}{\alpha_K} + \sum_{k=1}^K \frac{1}{\alpha_k k} \mathbb{E}\left[\lVert x_k \rVert  \right],
        \\
        &
        \mathbb{E}\left[\left\lVert \sum_{k=1}^K \gamma_k^{-1} (y_k - y_{k+1}) \right\rVert \right]
            \\ \leq &
        \frac{\mathbb{E} \left[\lVert y_1 \rVert \right]}{\gamma_1} + \frac{\mathbb{E}\left[\lVert y_{K+1} \rVert\right]}{\gamma_K} + \sum_{k=1}^K \frac{1}{\gamma_k k} \mathbb{E}\left[\lVert y_k \rVert \right].
    \end{split}
\end{equation*}
The expected errors of $x_k$ and $y_k$ for every $k \leq K$ are obtained from \Cref{lem:mse} and Jensen's inequality.
The asymptotic limit and the rate of convergence are then determined by \Cref{lem:CLT,lem:mse}, where we have that $\dwass \left(\sqrt{K} G \bar{x}_K , \bar{\pi}_x \right)$ is bounded by
\begin{equation}\label{eq:triangle_inequality_Wasserstein1}
\begin{split}
     \mathcal{O}\left(\frac{1}{K^{1/4}} \right) + \frac{1}{\sqrt{K}}\sum_{k=1}^K \frac{1}{\alpha_k k} \mathbb{E}\left[\lVert x_k \rVert   \right]
\end{split}
\end{equation}
and similarly for $\sqrt{K} \bar{y}_K$.
For the last term above, a more detailed version of \Cref{lem:mse} is used to identify singular behavior that arises for step size choices in \Cref{assumption:steps}, e.g., near the boundaries $a = 1/2$ and $b = a$.
Observing that the step sizes in the set $\mathcal{E}$ defined in \Cref{thm:clt} yield error rates faster than those outside of this set, we identify that the choice of $a = 1/2 + \mathcal{O}(1/\log K)$ and $b = a + \mathcal{O}(1/\log K)$ achieves the Wasserstein-1 rate $\mathcal{O}(K^{-1/4} + K^{-1/2}\log K)$, with leading order $K^{-1/4}$, while keeping the singular coefficients poly-logarithmic.

\Cref{cor:mae} is an immediate consequence of the dual form \eqref{eq:wasserstein_definition} of the Wasserstein-1 distance due to \cite{kantorovich1958space}.
The distance is defined as the supremum over all 1-Lipschitz functions, which we apply to the test function $h(x) = \lVert x \rVert$ to obtain
\begin{equation*}
    \begin{split}
        &\mathbb{E}\left[\lVert G \bar{x}_K\rVert \right]- \frac{1}{\sqrt{K}} \mathbb{E} \left[\lVert (G \bar{\Sigma}_{ff} G^T)^{1/2} Z_1 \rVert \right]
        \\ \leq
        &\frac{1}{\sqrt{K}} \dwass \left(\sqrt{K} G \bar{x}_K, (G \bar{\Sigma}_{ff} G^T)^{1/2} Z_1 \right)
        ,
        \\
        &\mathbb{E}\left[\lVert \Delta \bar{y}_K \rVert \right] - \frac{1}{\sqrt{K}} \mathbb{E} \left[\lVert (\Delta \bar{\Sigma}_{ss} \Delta^T)^{1/2} Z_2 \rVert \right]
        \\ \leq & \frac{1}{\sqrt{K}} \dwass \left(\sqrt{K} \Delta \bar{y}_K, (\Delta \bar{\Sigma}_{ss} \Delta^T)^{1/2} Z_2 \right)
        .
    \end{split}
\end{equation*}
The Wasserstein-1 distance is $o(1)$ for any step size satisfying \Cref{assumption:steps}, and an upper bound on $\mathbb{E}[\lVert G \bar{x}_K \rVert]$ and $\mathbb{E}[\lVert \Delta \bar{y}_K \rVert]$ is obtained.
The lower bound is obtained by using the test function $-\lVert x \rVert$.
\section{Conclusion}
We derived asymptotically tight finite-time bounds on the mean square error for two-time-scale stochastic approximation driven by martingale noise.
This was used to obtain finite-time bounds on the Wasserstein-1 distance between their Polyak-Ruppert averages and a Gaussian limit, which is a non-asymptotic central limit theorem for TSA-PR.
An important theoretical implication of our work is that it provides the first $\mathcal{O}(K^{-1/2})$ finite-time bound on the expected error, thereby closing the gap between convergence rates known for single- and two-time-scale algorithms.

A promising direction for future research is extending our analysis to the non-linear setting.
While finite-time mean-square error bounds for non-linear TSA have been established \cite{thinhFastNonlinear2024,chandak20261}, obtaining a non-asymptotic CLT for Polyak-Ruppert averages of non-linear TSA remains open.
\revised{The primary difficulty in the non-linear setting lies in establishing a suitable algebraic decomposition of the averaged iterates (akin to \eqref{eq:pr_expression_fast} and \eqref{eq:pr_expression_slow}) that clearly isolates a driving martingale from remainder terms.
The analytical technique in \cite{chandak20261} may be relevant to this goal: their approach provides tight convergence rates for the underlying non-averaged iterates, which is a necessary prerequisite for proving that such remainder terms decay sufficiently fast.}
We hope this work lays the groundwork for addressing the optimal finite time-scale gap for the non-linear setting.

\appendices
\crefalias{section}{appendix}

\section{Auxiliary Results}

\subsection{Usage of Assumptions \ref{assumption:first}--\ref{assumption:last}}\label{sec:recursion}
Here we state a property \postrevision{of the norm $\| \cdot \|_{P, \mathrm{F}}$} that will be used to prove \Cref{lem:mse,thm:clt}.
\postrevision{
If $-A$ is Hurwitz stable and $P \succ 0$ satisfies $A^TP+PA=I$, then $B=P^{1/2}AP^{-1/2}$ satisfies $B+B^T=P^{-1}$.
For a symmetric matrix $R$, we obtain
\begin{align*}
    &\lVert R-\alpha(AR+RA^T)\rVert_{P,\mathrm F}^2
    \\
    &=
    \Bigl\lVert
    P^{1/2}RP^{1/2}
    -\alpha\bigl(BP^{1/2}RP^{1/2}
    +P^{1/2}RP^{1/2}B^T\bigr)
    \Bigr\rVert_{\mathrm F}^2
    \\
    &\leq
    \lVert R\rVert_{P,\mathrm F}^2
    -2\alpha\operatorname{Tr}\!\left(
    \bigl(P^{1/2}RP^{1/2}\bigr)^2P^{-1}
    \right)
    +4\alpha^2\lVert B\rVert_{\mathrm{op}}^2
    \lVert R\rVert_{P,\mathrm F}^2
    \\
    &\leq
    \left(
    1-2\alpha\lambda_{\min}(P^{-1})
    +4\alpha^2\lVert B\rVert_{\mathrm{op}}^2
    \right)
    \lVert R\rVert_{P,\mathrm F}^2.
\end{align*}
Consequently, for some $\mu>0$ and all sufficiently small $\alpha>0$,
\begin{equation}\label{eq:symmetric_contraction}
    \lVert R-\alpha(AR+RA^T)\rVert_{P,\mathrm F}
    \leq
    \left(1-\alpha\mu\right)
    \lVert R\rVert_{P,\mathrm F}.
\end{equation}
\noindent
We apply \eqref{eq:symmetric_contraction} when $(A,P,\mu)$ is $(A_{ff},P_{ff},\mu_{ff}/2)$ or $(\Delta,P_\Delta,\mu_\Delta/2)$.
}

We also use that Assumption \ref{assumption:steps} implies $\alpha_k - \alpha_{k+1} \leq \frac{\alpha_{k+1}}{k} $, obtained from the elementary inequality $(\frac{k+1}{k})^a \leq (1+\frac{1}{k})$ for every $a \in (0, 1]$, and $\alpha_{k+1}^{-1} - \alpha_k^{-1} \leq (\alpha_{k} k)^{-1}$.
The same inequalities hold for the sequence $\{\gamma_k\}$.

\section{Mean Square Analysis of TSA}\label{app:tsa_mse}
In this section, we use the above properties to prove \Cref{lem:mse}.
The proof is divided into three parts, starting from the second moment of $\tilde{x}_k$, the joint covariance $\mathbb{E}[\tilde{x}_k \hat{y}_k^T]$, and the second moment of the slow-time-scale $\hat{y}_k$.
Only the second moments of $\tilde{x}_k$ and $\hat{y}_k$ are included in the statement of \Cref{lem:mse}, but the finite-time bound on the joint covariance is required as an intermediate step to establish the second moment of the slow-time-scale.

The transformed fast-time-scale $\tilde{x}_k$ is first analyzed because it is expressed recursively without an explicit dependence on $\hat{y}_k$.
This will be used to derive the finite-time bounds on the joint-time-scale covariance $\mathbb{E}[\tilde{x}_k \hat{y}_k^T]$, which is used as an intermediate result to derive the finite-time bound on the slow-time-scale covariance $\mathbb{E}[\hat{y}_k \hat{y}_k^T]$.

\postrevision{
To establish finite-time error bounds, our proofs repeatedly use the following consequence of \cite[Lemma 14]{kaledin2020finite}.
}
\begin{lemma}\label{lem:chung}
    \postrevision{
        Suppose $\{\alpha_k\}$ and $\{\gamma_k\}$ satisfy \Cref{assumption:steps}, and $(\eta_k,e_k)$ is one of
    \begin{equation*}
        \left(\alpha_k,c\gamma_k\right),\qquad
        \left(\alpha_k,c\frac{\gamma_k^2}{\alpha_k}\right),\qquad
        \left(\gamma_k,c\left(\frac{1}{k}+\frac{\gamma_k^2}{\alpha_k}\right)\right)
    \end{equation*}
    for some constant $c>0$.
    }
    \revised{Let $\{u_k\}_{k \geq 1}$ be a non-negative sequence satisfying the recursion
    \begin{equation*}
        u_{k+1} \leq (1 - \lambda \eta_k) u_k + \eta_k e_k ,
    \end{equation*}
    where $\lambda > 0$ is such that $\lambda \eta_k \in (0, 1)$.}
    Then, there exists a constant $M > 0$ such that for any $K \geq 1$,
    \begin{equation*}
        u_{K+1} \leq \prod_{k=1}^K (1 - \lambda \eta_k) u_1 + M e_K .
    \end{equation*}
\postrevision{The same rate in the conclusion holds when $e_k$ is upper bounded by any of the three listed rates.}
\end{lemma}

\subsection{Fast-Time-Scale Mean-Square Convergence}\label{sec:fast_mse}
Recall the notations $\hat{x}_K = x_K - x_\infty (y_K)$, $\hat{y}_K = y_K - y^*$, and $\tilde{L}_{k+1} \coloneqq L_{k+1} + A_{ff}^{-1} A_{fs}$.
Consider $\tilde{x}_k = \hat{x}_k + L_k \hat{y}_k$ described by the recursion
\begin{equation}\label{eq:fast_recursion}
    \tilde{x}_{k+1} = \tilde{x}_k - \alpha_k B_k^{ff} \tilde{x}_k + \alpha_k W_k + \gamma_k \tilde{L}_{k+1} V_k ,
\end{equation}
where
\begin{equation}\label{eq:Bt_ff}
    B_k^{ff} = A_{ff} + \frac{\gamma_k}{\alpha_k}\tilde{L}_{k+1} A_{sf} .
\end{equation}
We use $X_k$ to denote $\mathbb{E}[\tilde{x}_k \tilde{x}_k^T]$.
In addition,  let $\Sigma_{ff}$ be the matrix defined in \eqref{eq:covariances}.
Note that $\Sigma_{ff}$ exists since $-A_{ff}$ is Hurwitz.
Recall $\tilde{\delta}_x (k) = \lVert \mathbb{E}[\tilde{x}_k \tilde{x}_k^T] - \alpha_k \Sigma_{ff} \rVert_{P_{ff},\mathrm F}$.
\begin{lemma}\label{lem:fast_mse}
Under Assumptions \ref{assumption:first}-\ref{assumption:last}, there exists a problem-dependent constant $M_f > 0$ such that
    \begin{align*}
        \tilde{\delta}_x (K+1)
        &\leq
        \prod_{k=1}^K \left(1 - \alpha_k \frac{\mu_{ff}}{4}\right) \tilde{\delta}_x (1)
        \\ &\quad + M_f \mathrm{Err}_x (\alpha_K, \gamma_K, K)  ,
    \end{align*}
    where the rate $\mathrm{Err}_x (\alpha_K, \gamma_K, K)$ is given by
    \begin{equation}\label{eq:fast_rate}
        \mathrm{Err}_x (\alpha_K, \gamma_K, K)
        =
        \gamma_K \left(
    1 + \frac{\gamma_K}{\alpha_K} + \frac{1}{K \gamma_K}
    \right)
    \end{equation}
\end{lemma}

\begin{proof}

We start by evaluating the expectation of $\tilde{x}_k \tilde{x}_k^T$ from \eqref{eq:fast_recursion} conditioned on $\mathcal{H}_k$:
\begin{align*}
    &\mathbb{E} [\tilde{x}_{k+1} \tilde{x}_{k+1}^T | \mathcal{H}_k]
    \\= & \tilde{x}_k \tilde{x}_k^T - \alpha_k \left(B_k^{ff} \tilde{x}_k \tilde{x}_k^T + \tilde{x}_k \tilde{x}_k^T (B_k^{ff})^T - \alpha_k \Gamma_{ff} \right)
    \\
    +& \gamma_k^2 \tilde{L}_{k+1} \Gamma_{ss} \tilde{L}_{k+1}^T + \alpha_k \gamma_k (\Gamma_{fs} \tilde{L}_{k+1}^T + \tilde{L}_{k+1} \Gamma_{sf})
    \\ +& \alpha_k^2 B_k^{ff} \tilde{x}_k \tilde{x}_k^T (B_k^{ff})^T
    .
\end{align*}
Taking the unconditional expectations, we then obtain
\begin{align*}
    X_{k+1} &= X_k - \alpha_k (B_k^{ff} X_k + X_k (B_k^{ff})^T - \alpha_k \Gamma_{ff})
    \\ &
    + \gamma_k^2 \tilde{L}_{k+1} \Gamma_{ss} \tilde{L}_{k+1}^T + \alpha_k \gamma_k (\Gamma_{fs} \tilde{L}_{k+1}^T + \tilde{L}_{k+1} \Gamma_{sf})
    \\ &
    + \alpha_k^2 B_k^{ff} X_k (B_k^{ff})^T .
\end{align*}
Using the expression for $B_k^{ff}$ in \eqref{eq:Bt_ff}, the preceding equation is of the form
\begin{align*}
    X_{k+1} &= X_k - \alpha_k (A_{ff} X_k + X_k A_{ff}^T - \alpha_k \Gamma_{ff})
    \\ &\quad + \gamma_k R_k^{ff} (X_k) ,
\end{align*}
where
\begin{equation*}
\begin{split}
    R_k^{ff}  &\coloneqq
    - \left(\tilde{L}_{k+1} A_{sf} X_k + X_k A_{sf}^T \tilde{L}_{k+1}^T \right)
    + \frac{\alpha_k^2}{\gamma_k} B_k^{ff} X_k (B_k^{ff})^T
    \\ &+\left(
    \gamma_k \tilde{L}_{k+1} \Gamma_{ss} \tilde{L}_{k+1}^T
    +
     \alpha_k \Gamma_{fs} \tilde{L}_{k+1}^T
     +
     \alpha_k \tilde{L}_{k+1} \Gamma_{sf}\right)
     .
\end{split}
\end{equation*}
Using the identity for $\Gamma_{ff}$ in \eqref{eq:covariances} yields
\begin{align*}
    &X_{k+1} - \alpha_{k+1} \Sigma_{ff}
    \\ =& X_k - \alpha_k \Sigma_{ff}
    \\ -& \alpha_k \left(A_{ff} (X_k - \alpha_k \Sigma_{ff}) + (X_k - \alpha_k \Sigma_{ff}) A_{ff}^T\right)
    \\
    +& (\alpha_{k} - \alpha_{k+1}) \Sigma_{ff}
    + \gamma_k R_k^{ff}
    .
    \numberthis \label{eq:fast_equality}
\end{align*}
Using $\alpha_k^2/\gamma_k = \mathcal{O}(1)$ by \Cref{assumption:steps}, $\lVert L_k \rVert = \mathcal{O}(\gamma_k/\alpha_k)$, and the equivalence of matrix norms, we obtain for some constant $M_f' > 0$.
\begin{equation}\label{eq:fast_remainder}
    \postrevision{\lVert R_k^{ff}  \rVert_{P_{ff}, \mathrm F} \leq M_f' \left(\lVert X_k\rVert_{P_{ff}, \mathrm F} + \alpha_k + \gamma_k \right)}     .
\end{equation}
By the triangle inequality, we obtain
\begin{align*}
    \lVert R_k^{ff}  \rVert_{P_{ff},\mathrm F}
    &\leq M_f' \lVert X_k - \alpha_k \Sigma_{ff} \rVert_{P_{ff},\mathrm F}
    \\ &+
    M_f' \alpha_k \left(1 + \lVert \Sigma_{ff}\rVert_{P_{ff},\mathrm F}
    + \frac{\gamma_k}{\alpha_k} \right)
    .
\end{align*}
By \eqref{eq:symmetric_contraction} and $\alpha_k - \alpha_{k+1} \leq k^{-1} \alpha_{k+1}$, we have from \eqref{eq:fast_equality} the recursion
\begin{align*}
    &\tilde{\delta}_x (k+1)
    \\ & \leq \left(1 - \alpha_k \frac{\mu_{ff}}{2} + \gamma_k M_f'\right) \tilde{\delta}_x(k)
    + \frac{\alpha_{k+1}}{k} \lVert \Sigma_{ff} \rVert_{P_{ff},\mathrm F}
    \\ &
    + \alpha_k \gamma_k  M_f' \left(1 + \lVert \Sigma_{ff}  \rVert_{P_{ff},\mathrm F} + \frac{\gamma_k}{\alpha_k}\right) \leq
    \left(1 - \alpha_k \frac{\mu_{ff}}{4}\right) \tilde{\delta}_x (k)
    \\ &
    + \alpha_k \gamma_k M_f' \left(1 + \lVert \Sigma_{ff}  \rVert_{P_{ff},\mathrm F} + \frac{\gamma_k}{\alpha_k}+ \frac{\lVert \Sigma_{ff} \rVert_{P_{ff},\mathrm F}}{M_f'} \frac{1}{k \gamma_k} \right)
    .
\end{align*}
The preceding equation is of the form
\begin{equation*}
\begin{split}
    \tilde{\delta}_x (k+1) &\leq
    \left(1 - \alpha_k \frac{\mu_{ff}}{4}\right) \tilde{\delta}_x (k)
    + M_f^{''}  \alpha_k \mathrm{Err}_x(\alpha_k, \gamma_k, k)
    ,
\end{split}
\end{equation*}
for some constant $M_f^{''}$, where $\mathrm{Err}_x$ was defined in \eqref{eq:fast_rate}.
Applying \Cref{lem:chung} yields the stated bound for $\tilde{\delta}_x (K+1)$.
\end{proof}

\subsection{Joint-Time-Scale Mean Square Convergence}\label{sec:joint_mse}
The analysis of the slow iterate's covariance $\mathbb{E} [\hat{y}_k \hat{y}_k^T]$ involves an error term that depends on the deviation of cross covariance $C_k = \mathbb{E} [\tilde{x}_k \hat{y}_k^T]$ from $\gamma_k \Sigma_{fs}$.
Because $x, y$ can have different dimensions, the contraction is proved for the cross covariance in the norm
\begin{align*}
    \lVert C \rVert_{P_\Delta, P_{ff}}^2 = \sup_{v \neq 0} \frac{v^T C^T P_{ff} C v}{v^T P_\Delta v} .
\end{align*}
Recall the recursions in \eqref{eq:fast_slow_updates}.
Let $\Sigma_{fs}$ be the matrix defined in \eqref{eq:covariances} and define
\begin{equation*}
    \tilde{\delta}_{xy} (k) =
    \lVert C_{k} - \gamma_k \Sigma_{fs} \rVert_{P_\Delta, P_{ff}} .
\end{equation*}
This section is dedicated to proving a bound on $\tilde{\delta}_{xy}$, which will be used to establish a bound on $\delta_y(k)$ in \Cref{lem:mse}:
\begin{lemma}\label{lem:joint_mse}
    Under \Crefrange{assumption:first}{assumption:last}, there exists a problem-dependent constant $M_{fs} > 0$ such that
    \begin{equation*}
    \begin{split}
    \tilde{\delta}_{xy} (K+1)
        &\leq
    \prod_{k=1}^K \left(1 - \alpha_k \frac{\mu_{ff}}{4}\right) \tilde{\delta}_{xy}(1)
        \\ &\quad + M_{fs}     \mathrm{Err}_{xy} (\alpha_K, \gamma_K, K),
    \end{split}
    \end{equation*}
    where the rate $\mathrm{Err}_{xy}(\alpha_K, \gamma_K, K)$ is defined as
    \begin{equation}\label{eq:Errxy}
    \mathrm{Err}_{xy}(\alpha_K, \gamma_K, K)
    =
        \frac{\gamma_K}{\alpha_K}  \left(\frac{1}{K} +
      (\alpha_K + \gamma_K)^2
     + \gamma_K \right)
     .
    \end{equation}
\end{lemma}

\begin{proof}
We first obtain a recursion for $C_k = \mathbb{E}[\tilde{x}_k \hat{y}_k^T]$ by taking the outer product between the recursion for $\tilde{x}_{k+1}$ and $\hat{y}_{k+1}$ in \eqref{eq:fast_slow_updates}.
Grouping terms by their rates (i.e., order-dependency on $t$ as determined by the step sizes), we obtain
\begin{align*}
    C_{k+1}
    &= C_k - \gamma_k \left( \frac{\alpha_k}{\gamma_k} A_{ff} C_k + X_k A_{sf}^T - \alpha_k \Gamma_{fs}
    \right)
    \\ &
    - \gamma_k \tilde{L}_{k+1} \left( A_{sf} C_k - \gamma_k \Gamma_{ss}\right)
    - \gamma_k C_k \Delta^T
    \\ &
    + \alpha_k \gamma_k A_{ff} X_k A_{sf}^T
    \\ &
    +
    \gamma_k \left(\alpha_k A_{ff} C_k \Delta^T + \gamma_k \tilde{L}_{k+1} A_{sf} X_k A_{sf}^T\right)
    \\ &
    - \gamma_k \left(\alpha_k A_{ff} C_k L_k^T A_{sf}^T - \gamma_k \tilde{L}_{k+1} A_{sf} C_k \Delta^T\right)
    \\ &
    + \gamma_k C_k L_k^T A_{sf}^T
    - \gamma_k^2 \tilde{L}_{k+1} A_{sf} C_k L_k^T A_{sf}^T
    .
\end{align*}
This recursion is of the form
\begin{equation}\label{eq:joint_covariance_recursion}
    C_{k+1} = C_k - \alpha_k A_{ff} C_k - \gamma_k (X_k A_{sf}^T - \alpha_k \Gamma_{fs}) + \gamma_k R_k^{fs} ,
\end{equation}
where
\begin{equation}\label{eq:def_Rfs}
\begin{split}
R_k^{fs} &=
    -  C_k \Delta^T
    + \alpha_k  A_{ff} C_k \Delta^T
    +  C_k L_k^T A_{sf}^T
    \\
    &
    - \alpha_k  A_{ff}     \left(C_k L_k^T A_{sf}^T - X_k A_{sf}^T \right)
    \\ &-  \tilde{L}_{k+1} A_{sf} C_k
    \\ &+ \gamma_k \tilde{L}_{k+1} A_{sf} \left(C_k \Delta^T - C_k L_k^T A_{sf}^T + X_k A_{sf}^T\right)
    \\ &
    + \gamma_k \tilde{L}_{k+1} \Gamma_{ss}
    .
\end{split}
\end{equation}
Substituting \eqref{eq:covariances} for $\Gamma_{fs}$ in \eqref{eq:joint_covariance_recursion},
\begin{align*}
    C_{k+1} &
    = C_k - \alpha_k A_{ff} C_k
    \\ &- \gamma_k \left(
        (X_k - \alpha_k \Sigma_{ff}) A_{sf}^T - \alpha_k A_{ff} \Sigma_{fs}
    \right)
    + \gamma_k R_k^{fs}
    \\ &
    = C_k - \alpha_k A_{ff}( C_k - \gamma_k \Sigma_{fs})
    \\ &
    - \gamma_k \left(X_k - \alpha_k \Sigma_{ff}\right) A_{sf}^T
    + \gamma_k R_k^{fs}
    .
\end{align*}
Subtracting $\gamma_{k+1} \Sigma_{fs}$ on both sides,
\begin{equation}\label{eq:joint_recursion}
\begin{split}
    C_{k+1} - \gamma_{k+1} \Sigma_{fs}
    &=
    (I - \alpha_k A_{ff}) (C_k - \gamma_k \Sigma_{fs})
    \\ &
    + (\gamma_k - \gamma_{k+1}) \Sigma_{fs}
    \\ &
    - \gamma_k (X_k - \alpha_k \Sigma_{ff} )A_{sf}^T + \gamma_k R_k^{fs} .
\end{split}
\end{equation}
Applying the operator norm to $R_k^{fs}$ in \eqref{eq:def_Rfs} and using that $\tilde{L}_{k+1}$ is uniformly bounded, we obtain for some constant $M_{fs}'$
\begin{equation}\label{eq:cross_remainder}
\begin{split}
    \lVert R_k^{fs} \rVert_{P_\Delta, P_{ff}}
    &\leq M_{fs}'\left(
        \lVert C_k \rVert_{P_\Delta, P_{ff}}
    + \left(\alpha_k + \gamma_k\right) \lVert X_k \rVert_{P_{ff},\mathrm F}
    + \gamma_k
    \right).
\end{split}
\end{equation}
By the triangle inequalities
\begin{align*}
    \lVert C_k \rVert_{P_\Delta, P_{ff}}
    &\leq \lVert C_k - \gamma_k \Sigma_{fs} \rVert_{P_{\Delta}, P_{ff}} + \gamma_k \lVert \Sigma_{fs} \rVert_{P_\Delta, P_{ff}},
    \\
    \lVert X_k \rVert_{P_{ff},\mathrm F} &\leq \lVert X_k - \alpha_k \Sigma_{ff} \rVert_{P_{ff},\mathrm F} + \alpha_k \lVert \Sigma_{ff} \rVert_{P_{ff},\mathrm F} ,
\end{align*}
we have
\begin{align*}
    &\lVert R_k^{fs} \rVert_{P_\Delta, P_{ff}}
    \\ \leq &
    M_{fs}'
    \Big(\lVert C_k - \gamma_k \Sigma_{fs} \rVert_{P_\Delta, P_{ff}}
    \\ &
    +(\alpha_k + \gamma_k) \lVert X_k - \alpha_k \Sigma_{ff} \rVert_{P_{ff},\mathrm F}
    \\
    + &\gamma_k (\lVert \Sigma_{fs} \rVert_{P_\Delta, P_{ff}} +1 )+ \alpha_k (\alpha_k + \gamma_k) \Big)
    \\
    \leq &
    M_{fs}' \Big(\lVert C_k - \gamma_k \Sigma_{fs} \rVert_{P_\Delta, P_{ff}} + (\alpha_k + \gamma_k)^2
    \\ & + \gamma_k (\lVert \Sigma_{fs} \rVert_{P_{\Delta}, P_{ff}} + 1) \Big)
\end{align*}
where the last line used $\lVert X_k - \alpha_k \Sigma_{ff}\rVert_{P_{ff},\mathrm F} = \mathcal{O}(\gamma_k)$ from \Cref{lem:fast_mse}.
Continuing from \eqref{eq:joint_recursion}, we have the contraction
\begin{align*}
    &\lVert C_{k+1} - \gamma_{k+1} \Sigma_{fs}\rVert_{P_{\Delta}, P_{ff}}
    \\ &\leq
    \left(1 - \alpha_k \frac{\mu_{ff}}{2} + \gamma_k M_{fs}'\right) \lVert C_k - \gamma_k \Sigma_{fs} \rVert_{P_{\Delta}, P_{ff}}
    \\ &
    + \frac{\gamma_k}{k} \lVert \Sigma_{fs} \rVert_{P_{\Delta}, P_{ff}}
    + \gamma_k \lVert X_k - \alpha_k \Sigma_{ff}\rVert_{P_{ff},\mathrm F} \lVert A_{sf}^T\rVert_{P_{\Delta}, P_{ff}}
    \\ &+
    M_{fs}' \gamma_k \left( (\alpha_k + \gamma_k)^2 + \gamma_k (\lVert \Sigma_{fs} \rVert_{P_\Delta, P_{ff}} + 1) \right)
    \\ &
    \leq \left(1 - \alpha_k \frac{\mu_{ff}}{4}\right) \lVert C_k - \gamma_k \Sigma_{fs}\rVert_{P_{\Delta}, P_{ff}}
    \\ &
    +
    M_{fs}' \Big(\frac{\gamma_k}{k} \lVert \Sigma_{fs} \rVert_{P_\Delta, P_{ff}} +
     \gamma_k (\alpha_k + \gamma_k)^2
     \\ &
     + \gamma_k^2 (\lVert A_{sf}^T \rVert_{P_\Delta, P_{ff}} + \lVert \Sigma_{fs} \rVert_{P_\Delta, P_{ff}} + 1) \Big)
     .
\end{align*}
The above recursion is of the form, for some constant $M_{fs}''$,
\begin{align*}
    \tilde{\delta}_{xy}(k+1)
        &\leq \left(1 - \alpha_k \frac{\mu_{ff}}{4}\right) \tilde{\delta}_{xy}(k)
        \\ &\quad + \alpha_k  M_{fs}'' \mathrm{Err}_{xy} (\alpha_k, \gamma_k, k),
\end{align*}
with $\mathrm{Err}_{xy}$ in \eqref{eq:Errxy}.
Applying \Cref{lem:chung} yields the stated bound for $\tilde{\delta}_{xy}(K+1)$.
\end{proof}

\subsection{Slow-Time-Scale Mean Square Convergence}\label{sec:slow_mse}
The bounds above for fast- and cross-covariance matrices $X_{K}$ and $C_{K}$ are used to derive the convergence rate of $Y_K = \mathbb{E}[\hat{y}_K \hat{y}_K^T]$.
Recall $\delta_y(k) = \lVert Y_k - \gamma_k \Sigma_{ss} \rVert_{P_\Delta, \mathrm F}$ defined in \eqref{eq:def_delta_y}.
\begin{lemma}\label{lem:slow_mse}
    Let $\Sigma_{ss} \succ 0$ be the matrix defined in \eqref{eq:covariances}.
    Under \Crefrange{assumption:first}{assumption:last}, there exists a problem-dependent constant $M_s > 0$ such that
    \begin{align*}
    \delta_y (K+1)
    &\leq
    \prod_{k=1}^K \left(1 - \gamma_k \frac{\mu_\Delta}{4}\right) \delta_y (1)
    \\ &\quad +
    M_s \mathrm{Err}_y (\alpha_K, \gamma_K, K),
    \end{align*}
    where the rate $\mathrm{Err}_y (\alpha_K, \gamma_K, K)$ is given by
    \begin{equation}\label{eq:Erry}
        \mathrm{Err}_y (\alpha_K, \gamma_K, K)
        = \frac{1}{K} + \frac{\gamma_K^2}{\alpha_K}
        .
    \end{equation}
\end{lemma}
\begin{proof}
Recall the slow iterate's recursion
\begin{equation*}
    \hat{y}_{k+1} = (I - \gamma_k \Delta) \hat{y}_k + \gamma_k A_{sf} L_k \hat{y}_k - \gamma_k A_{sf} \tilde{x}_k + \gamma_k V_k .
\end{equation*}
By direct substitution, we obtain a recursion for $\mathbb{E}[\hat{y}_{k+1} \hat{y}_{k+1}^T]$.
Grouping terms by magnitude, we obtain
\begin{equation}\label{eq:slow_covariance_recursion}
\begin{split}
    Y_{k+1} &= Y_k
    - \gamma_k \Big(\Delta Y_k + Y_k \Delta^T
    \\ &\quad + (C_k^T A_{sf}^T + A_{sf} C_k) - \gamma_k \Gamma_{ss}  \Big)
    + \frac{\gamma_k^2}{\alpha_k} R_k^{ss}  ,
\end{split}
\end{equation}
where the remainder $R_k^{ss}$ is defined as
\begin{align*}
    R_k^{ss}
    &=
    \alpha_k (\Delta Y_k \Delta^T)
    \\ &+ \frac{\alpha_k}{\gamma_k} \left((I - \gamma_k \Delta) Y_k L_k^T A_{sf}^T + ((I - \gamma_k \Delta) Y_k L_k^T A_{sf}^T)^T\right)
    \\
    &+ \alpha_k \left( \Delta C_k^T A_{sf}^T +  A_{sf} C_k \Delta^T \right) + \alpha_k A_{sf} L_k Y_k L_k^T A_{sf}^T
    \\&- \alpha_k \left(A_{sf} L_k C_k^T A_{sf}^T + (A_{sf} L_k C_k^T A_{sf}^T)^T\right)
    \\
    &+ \alpha_k A_{sf} X_k A_{sf}^T .
\end{align*}
Using that $\lVert L_k \rVert = \mathcal{O}(\gamma_k/\alpha_k)$, we have for some constants $M_s', M_s'' > 0$
\begin{align*}
    &\lVert R_k^{ss}  \rVert_{P_{\Delta},\mathrm F}
    \\ \leq &
    M_s' \Big(\lVert Y_k \rVert_{P_{\Delta},\mathrm F}
    + (\alpha_k + \gamma_k) \lVert A_{sf} C_k \rVert_{P_{\Delta}}
    \\ &
    + \alpha_k \lVert A_{sf} X_k A_{sf}^T \rVert_{P_{\Delta},\mathrm F}
    \Big)
    \\
     \leq &
    M_s' \left(\lVert Y_k - \gamma_k \Sigma_{ss} \rVert_{P_\Delta,\mathrm F} + \gamma_k \lVert \Sigma_{ss} \rVert_{P_\Delta,\mathrm F}\right)
    \\
     +& M_s' (\alpha_k + \gamma_k) \left(\lVert A_{sf} (C_k - \gamma_k \Sigma_{fs}) \rVert_{P_\Delta} + \gamma_k \lVert A_{sf} \Sigma_{fs} \rVert_{P_\Delta}\right)
    \\  + &
    M_s' \alpha_k \left(\lVert A_{sf} (X_k - \alpha_k \Sigma_{ff}) A_{sf}^T \rVert_{P_\Delta,\mathrm F}
    + \alpha_k \lVert A_{sf} \Sigma_{ff} A_{sf}^T \rVert_{P_\Delta,\mathrm F} \right)
    \\
    \leq &
     M_s' \delta_y(k)
    + M_s'' (\alpha_k + \gamma_k) \tilde{\delta}_{xy}(k)
    + M_s'' \alpha_k \tilde{\delta}_x (k)
    \\
    + & M_s'' \gamma_k
    .
\end{align*}
Moreover, bounds on $\tilde{\delta}_{x} (k)$ and $\tilde{\delta}_{xy}(k)$ have been established in \Cref{lem:fast_mse,lem:joint_mse}.
Substituting the expression for $\Gamma_{ss}$ in \eqref{eq:covariances}, \eqref{eq:slow_covariance_recursion} is expressed as
\begin{align*}
    &
    Y_{k+1} \\= & Y_k
    - \gamma_k \Big(
        \Delta Y_k + Y_k \Delta^T
        \\ &\quad + \gamma_k (\Sigma_{fs}^T A_{sf}^T + A_{sf} \Sigma_{fs}) - \gamma_k \Gamma_{ss}
    \Big)
    \\
    -& \gamma_k \left((C_k - \gamma_k \Sigma_{fs})^T A_{sf}^T + A_{sf} (C_k - \gamma_k \Sigma_{fs})\right)
    + \frac{\gamma_k^2}{\alpha_k} R_k^{ss}
    \\
    =& Y_k
    - \gamma_k \left(\Delta (Y_k - \gamma_k \Sigma_{ss}) + (Y_k - \gamma_k \Sigma_{ss})\Delta^T \right)
    \\
    - &\gamma_k \left((C_k - \gamma_k \Sigma_{fs})^T A_{sf}^T + A_{sf} (C_k - \gamma_k \Sigma_{fs})\right)
    + \frac{\gamma_k^2}{\alpha_k} R_k^{ss}  .
\end{align*}
Subtracting both sides by $\gamma_{k+1} \Sigma_{ss}$,
\begin{align*}
    &Y_{k+1} - \gamma_{k+1} \Sigma_{ss}
    \\=& Y_k - \gamma_k \Sigma_{ss}
    \\ & - \gamma_k \left(\Delta (Y_k - \gamma_k \Sigma_{ss}) + (Y_k - \gamma_k \Sigma_{ss})\Delta^T \right)
    \\
    +& (\gamma_k - \gamma_{k+1}) \Sigma_{ss}
    \\
    -& \gamma_k \left((C_k - \gamma_k \Sigma_{fs})^T A_{sf}^T + A_{sf} (C_k - \gamma_k \Sigma_{fs})\right)
    \\ & + \frac{\gamma_k^2}{\alpha_k} R_k^{ss} (Y_k)
    .
\end{align*}
Using \eqref{eq:symmetric_contraction} and $\gamma_k - \gamma_{k+1} \leq \gamma_k/k$, we obtain by taking the weighted norm $\lVert \cdot \rVert_{P_\Delta, \mathrm F}$ on both sides
\begin{align*}
    &\delta_y(k+1) \leq
    \left(1 - \gamma_k \frac{\mu_\Delta}{2} + \frac{\gamma_k^2}{\alpha_k} M_s' \right) \delta_y(k)
    + \frac{\gamma_k}{k} \lVert \Sigma_{ss} \rVert_{P_{\Delta},\mathrm F}
    \\ &
    + \gamma_k \left\lVert (C_k - \gamma_k \Sigma_{fs})^T A_{sf}^T + A_{sf} (C_k - \gamma_k \Sigma_{fs})\right\rVert_{P_{\Delta},\mathrm F}
    \\ &
    + \frac{\gamma_k^3}{\alpha_k} M_s'' \left(
    (\alpha_k + \gamma_k) \frac{\gamma_k}{\alpha_k} + \alpha_k
    + 1 + \gamma_k (\alpha_k + \gamma_k) + \frac{\alpha_k^2}{\gamma_k}
    \right)
    \\ &
    \leq
    \left(1 - \gamma_k \frac{\mu_\Delta}{4} \right) \delta_y(k)
    +
    M_s''' \frac{\gamma_k}{k}
    \\ &
    + M_s''' \frac{\gamma_k^3}{\alpha_k}\left(
    (\alpha_k + \gamma_k) \frac{\gamma_k}{\alpha_k} + \alpha_k
    + 1 + \gamma_k (\alpha_k + \gamma_k) + \frac{\alpha_k^2}{\gamma_k} \right)
    .
\end{align*}
Using that the above equation is of the form
\begin{align*}
    \delta_y(k+1)
    &\leq \left(1 - \gamma_k \frac{\mu_\Delta}{4}\right) \delta_y(k)
    + M_s'''' \gamma_k \mathrm{Err}_y (\alpha_k, \gamma_k, k),
\end{align*}
where the rate $\mathrm{Err}_y (\alpha_k, \gamma_k, k)$ was defined in \eqref{eq:Erry}.
Applying \Cref{lem:chung} yields the stated bound for $\delta_{y}(K+1)$.
\end{proof}

\section{Non-Asymptotic CLT for TSA-PR}\label{app:clt}
A finite-time bound on the Wasserstein-1 distance between the optimality gaps achieved by TSA-PR and their limiting Gaussian distributions is established.

\subsection{Proof of Theorem \ref{thm:clt}}\label{proof:clt}
\textbf{Step 1:} We write the recursions \eqref{eq:ttsa} in the form
\begin{align*}
    G x_k &= (W_k - A_{fs} A_{ss}^{-1} V_k) + \alpha_k^{-1} (x_k - x_{k+1})
    \\ &- \gamma_k^{-1} A_{fs} A_{ss}^{-1} (y_k - y_{k+1})
    \\
    \Delta y_k &= (V_k - A_{sf} A_{ff}^{-1} W_k) + \gamma_k^{-1} (y_k - y_{k+1})
    \\ &
    - A_{sf} A_{ff}^{-1} \alpha_k^{-1} (x_k - x_{k+1}) ,
\end{align*}
where $G, \Delta$ are the Schur complements
\begin{align*}
    G = A_{ff} - A_{fs} A_{ss}^{-1} A_{sf} , \quad
    \Delta = A_{ss} - A_{sf} A_{ff}^{-1} A_{fs} .
\end{align*}
\begin{proof}
    Let us start with the fast-time-scale variable update
    \begin{align*}
        x_{k+1} &= x_k - \alpha_k (A_{ff} x_k + A_{fs} y_k - W_k)
        \\
        \Rightarrow x_k &= (\alpha_k A_{ff})^{-1} (x_{k} - x_{k+1}) - A_{ff}^{-1} (A_{fs} y_k - W_k) .
    \end{align*}
    Substituting into the slow-time-scale variable's recursion,
    \begin{align*}
        y_{k+1} &= y_k - \gamma_k \left(
            A_{sf} x_k + A_{ss} y_k - V_k
        \right)
        \\
        &= y_k - \gamma_k A_{sf} A_{ff}^{-1} \alpha_k^{-1} (x_{k} - x_{k+1})
        \\ &+ \gamma_k A_{sf} A_{ff}^{-1} A_{fs} y_k - \gamma_k A_{sf} A_{ff}^{-1} W_k
        \\ & - \gamma_k A_{ss} y_k + \gamma_k V_k .
    \end{align*}
    Using $\Delta = A_{ss} - A_{sf} A_{ff}^{-1} A_{fs}$, we have
    \begin{align*}
         y_{k+1}
         &= (I - \gamma_k \Delta) y_k - \frac{\gamma_k}{\alpha_k} A_{sf} A_{ff}^{-1} (x_{k} - x_{k+1})
         \\ &+ \gamma_k (V_k - A_{sf} A_{ff}^{-1} W_k) ,
    \end{align*}
    which is rearranged as
    \begin{align*}
         \gamma_k \Delta y_k
         &=
         y_k - y_{k+1} - \frac{\gamma_k}{\alpha_k} A_{sf} A_{ff}^{-1} (x_{k} - x_{k+1}) \\ &
         + \gamma_k (V_k - A_{sf} A_{ff}^{-1} W_k).
    \end{align*}
    Dividing both sides by the step size $\gamma_k$, we have
    \begin{align*}
        \Delta y_k &= \gamma_k^{-1} (y_k - y_{k+1}) - \alpha_k^{-1} A_{sf} A_{ff}^{-1} (x_{k} - x_{k+1})
        \\ &
        +(V_k - A_{sf} A_{ff}^{-1} W_k) .
    \end{align*}
    We repeat the same steps for the fast iterate:
    Using that $A_{ss} y_k = \gamma_k^{-1} (y_k - y_{k+1}) - (A_{sf} x_k - V_k)$, we have by substitution
    \begin{align*}
        x_{k+1} &= x_k - \alpha_k (A_{ff} x_k - W_k) - \alpha_k A_{fs} y_k
        \\
        &= x_k - \alpha_k A_{ff} x_k + \alpha_k W_k
        \\ &- \alpha_k A_{fs} A_{ss}^{-1}\left(\gamma_k^{-1} (y_k - y_{k+1}) - (A_{sf} x_k - V_k) \right)
        \\ & =
        x_k - \alpha_k \left(A_{ff} - A_{fs} A_{ss}^{-1} A_{sf} \right) x_k
        \\ &+ \alpha_k (W_k - A_{fs} A_{ss}^{-1} V_k) - \frac{\alpha_k}{\gamma_k} A_{fs} A_{ss}^{-1} (y_k - y_{k+1})
        \\ &
        = x_k - \alpha_k G x_k
        \\ &+ \alpha_k (W_k - A_{fs} A_{ss}^{-1} V_k) - \frac{\alpha_k}{\gamma_k} A_{fs} A_{ss}^{-1} (y_k - y_{k+1})
        .
    \end{align*}
    Rearranging, we have
    \begin{align*}
        G x_k &= \alpha_k^{-1} (x_k - x_{k+1}) + (W_k - A_{fs} A_{ss}^{-1} V_k) \\ &- \gamma_k^{-1} A_{fs} A_{ss}^{-1} (y_k - y_{k+1}) .
    \end{align*}
\end{proof}
The expressions in \eqref{eq:pr_expression_fast} and \eqref{eq:pr_expression_slow} are obtained by averaging over $k \in \{1, \cdots, K\}$.

\textbf{Step 2:}
We obtain non-asymptotic bounds on $\dwass (\sqrt{K} G(\bar{x}_K - x^*), \bar{\pi}_x)$ and $\dwass (\sqrt{K}\Delta(\bar{y}_K - y^*), \bar{\pi}_y)$.
The (asymptotic) convergence to Gaussian limits can be proved by showing that (1) the first noise terms above satisfy Lindeberg's condition and that (2) the remaining terms converge to zero with probability 1.
Slutsky's theorem then implies that the sum $N_k + E_k \to N + c$ when the noise $N_k$ converges in distribution to a random variable $N$ and the error terms $E_k$ converge to a constant $c$ with probability 1.

For the non-asymptotic analysis, we use \Cref{lem:CLT} to obtain the bound as in \eqref{eq:triangle_inequality_Wasserstein1}.
Next, we apply Jensen's inequality with the error bounds in \Cref{lem:mse} to obtain a bound on $\mathbb{E}[\lVert x_k \rVert]$ and $\mathbb{E}[\lVert y_k \rVert]$ for all $k$ when $x^*, y^*$ are zero (due to zero bias assumed for simplicity).
Observe the identities
\begin{align*}
    \sum_{k=1}^K \alpha_k^{-1} (x_k - x_{k+1})
    &= \frac{x_1}{\alpha_1} - \frac{x_{K+1}}{\alpha_{K}} + \sum_{k=1}^{K-1} (\alpha_{k+1}^{-1} - \alpha_{k}^{-1}) x_{k+1}
    \\
    \sum_{k=1}^K \gamma_k^{-1}(y_k - y_{k+1}) &= \frac{y_1}{\gamma_1} - \frac{y_{K+1}}{\gamma_{K}} + \sum_{k=1}^{K-1} (\gamma_{k+1}^{-1} - \gamma_k^{-1})y_{k+1}
    .
\end{align*}
The step sizes in \Cref{assumption:steps} satisfy
\begin{align*}
    \alpha_{k+1}^{-1} - \alpha_{k}^{-1} \leq \frac{1}{k\alpha_k},
    \quad
    \gamma_{k+1}^{-1} - \gamma_k^{-1} \leq \frac{1}{k \gamma_k}
    ,
\end{align*}
and we use the triangle inequality to obtain the bounds
\begin{align*}
    &\mathbb{E} \left[ \lVert \sum_{k=1}^K \alpha_k^{-1} (x_k - x_{k+1}) \rVert \right]
    \\
    \leq
    & \alpha_1^{-1} \mathbb{E}\left[\lVert x_1 \rVert\right]  + \alpha_K^{-1} \mathbb{E}\left[\lVert x_{K+1} \rVert\right]
    + (\alpha_1)^{-1} \sum_{k=1}^K k^{a - 1} \mathbb{E}\left[ \lVert  x_k \rVert     \right]
\end{align*}
and
\begin{align*}
    &\mathbb{E} \left[\lVert \sum_{k=1}^K \gamma_k^{-1} (y_k - y_{k+1}) \rVert \right]
    \\ \leq &\gamma_1^{-1} \mathbb{E}\left[ \lVert y_1 \rVert \right] + \gamma_K^{-1} \mathbb{E}\left[ \lVert y_{K+1} \rVert\right] + (\gamma_1)^{-1} \sum_{k=1}^K k^{b - 1} \mathbb{E}\left[ \lVert  y_k \rVert \right].
\end{align*}
The expected norm can be bounded using
\begin{equation}\label{eq:yhat_appendix}
    \mathbb{E}\left[\lVert y_{k+1} \rVert \right]
    \leq \sqrt{\mathrm{Tr} \mathbb{E} \left[y_{k+1} y_{k+1}^T \right]}
    \leq \sqrt{d_s \lVert \mathbb{E}\left[\hat{y}_{k+1} \hat{y}_{k+1}^T \right]\rVert}
    ,
\end{equation}
where the first step uses Jensen's inequality and the second step uses that $\mathbb{E} [y_{k+1} y_{k+1}^T]$ has non-negative eigenvalues.
Similarly, we have a bound on $\lVert \mathbb{E} [\tilde{x}_k \tilde{x}_k^T ]\rVert$ using \Cref{lem:fast_mse}:
\begin{align*}
    &\mathbb{E}\left[\lVert \hat{x}_{k+1} \rVert \right]
    \\ \leq &
    \sqrt{d_{f} \lVert \mathbb{E}\left[\hat{x}_{k+1} \hat{x}_{k+1}^T \right]\rVert }
    \\ \leq &
    \sqrt{d_f} \Big(\lVert \mathbb{E}\left[\hat{x}_{k+1} \hat{x}_{k+1}^T\right] - \mathbb{E} \left[\tilde{x}_{k+1} \tilde{x}_{k+1}^T\right] \rVert
    \\ &\quad + \lVert \mathbb{E}\left[\tilde{x}_{k+1} \tilde{x}_{k+1}^T\right] \rVert\Big)^{1/2}.
\end{align*}
Using that
\begin{align*}
    \mathbb{E}\left[\tilde{x}_{k+1} \tilde{x}_{k+1}^T \right]
    &=
    \mathbb{E} \left[\hat{x}_{k+1} \hat{x}_{k+1}^T \right]
    \\ &
    + \left(\mathbb{E} \left[\hat{x}_{k+1} \hat{y}_{k+1}^T \right] L_{k+1}^T
    + L_{k+1} \mathbb{E} \left[\hat{y}_{k+1} \hat{x}_{k+1}^T\right]
    \right)
    \\ &
    + L_{k+1} \mathbb{E} \left[\hat{y}_{k+1} \hat{y}_{k+1}^T \right]L_{k+1}^T,
\end{align*}
we obtain the bound
\begin{equation}\label{eq:xhat_appendix}
\begin{split}
    \mathbb{E}\left[\lVert \hat{x}_{k+1} \rVert \right]
    &\leq
    \sqrt{d_f} \Big(
    2 \lVert L_{k+1} \rVert \lVert \mathbb{E} \left[\hat{x}_{k+1} \hat{y}_{k+1}^T \right] \rVert
    \\ &
    + \lVert L_{k+1} \rVert^2  \left\lVert \mathbb{E} \left[\hat{y}_{k+1} \hat{y}_{k+1}^T \right] \right\rVert
    \\ &
    + \left\lVert \mathbb{E} \left[\tilde{x}_{k+1} \tilde{x}_{k+1}^T \right] \right\rVert
    \Big)^{1/2}
    .
\end{split}
\end{equation}
Omitting the $\sqrt{d_f}$ dependency and multiplicative constants from \eqref{eq:xhat_appendix}, we obtain that the error rate of $\mathbb{E}[\lVert \hat{x}_{k+1} \rVert]$ is bounded by
\begin{align*}
    &\sqrt{2\lVert L_{k+1} \rVert} \sqrt{
     \lVert \mathbb{E} \left[\tilde{x}_{k+1} \hat{y}_{k+1}^T\right] \rVert
     + \lVert L_{k+1} \rVert  \lVert \mathbb{E} \left[\hat{y}_{k+1} \hat{y}_{k+1}^T \right]\rVert
    }
    \\ +& \sqrt{\alpha_k \lVert \Sigma_{ff} \rVert +  \mathrm{Err}_x (\alpha_k, \gamma_k, k)}
    \\ \leq &
    \sqrt{\frac{\gamma_k^2}{\alpha_k}}
    + \frac{\gamma_k^{3/2}}{\alpha_k} +
    \sqrt{\frac{\gamma_k}{\alpha_k} \mathrm{Err}_{x y} (\alpha_k, \gamma_k, k)}
    \\ +& \frac{\gamma_k}{\alpha_k} \sqrt{\mathrm{Err}_y (\alpha_k, \gamma_k, k) }
    + \sqrt{\alpha_k}
    + \sqrt{\mathrm{Err}_x (\alpha_k, \gamma_k, k)}
    .
\end{align*}
For sufficiently large $k$, the geometrically decaying terms in \Cref{lem:slow_mse} become negligible. 
Applying this to \eqref{eq:yhat_appendix} and \eqref{eq:xhat_appendix} alongside $\|L_k\| = \mathcal{O}(\gamma_k / \alpha_k)$, we obtain for some constants $M_1$ and $M_2$,

\begin{align*}
    \mathbb{E}\left[\lVert y_{k+1} \rVert \right]
    &\leq
    M_2 \left( \sqrt{\gamma_k} +  \sqrt{\mathrm{Err}_y (\alpha_k, \gamma_k, k)} \right)
    \\
    \mathbb{E}\left[\lVert \hat{x}_{k+1} \rVert \right]
     &\leq
    M_1 \Big(\sqrt{\alpha_k} + \sqrt{\frac{\gamma_k}{\alpha_k} \mathrm{Err}_{x y} (\alpha_k, \gamma_k, k)}
    \\ &+ \frac{\gamma_k}{\alpha_k} \sqrt{\mathrm{Err}_y (\alpha_k, \gamma_k, k) }
    + \sqrt{\mathrm{Err}_x (\alpha_k, \gamma_k, k)}
    \Big) .
\end{align*}
\textbf{Step 3:} Optimal step size and dominant terms.
We next analyze the sum
\begin{equation}\label{eq:W1_bound}
\begin{split}
    &\frac{1}{\sqrt{K}} \sum_{k=1}^K \left(k^{a - 1} \mathbb{E} \left[\lVert x_k \rVert\right] + k^{b-1} \mathbb{E}\left[\lVert y_k \rVert\right] \right)
    \\ & \leq
    \frac{M_1}{\sqrt{K}} \sum_{k=1}^K \left(k^{a/2 - 1} + k^{b/2 - 1} \right)
    \\ &+
    \underbrace{\frac{M_1}{\sqrt{K}} \sum_{k=1}^K  k^{a-1} \sqrt{\mathrm{Err}_x (\alpha_k, \gamma_k, k)}}_{(a)}
    \\ &
    +
    \underbrace{\frac{M_1}{\sqrt{K}} \sum_{k=1}^K k^{a-1} \sqrt{\frac{\gamma_k}{\alpha_k} \mathrm{Err}_{xy} (\alpha_k, \gamma_k, k)}}_{(b)}
    \\ &
    +
    \underbrace{\frac{M_1}{\sqrt{K}} \sum_{k=1}^K k^{a-1} \frac{\gamma_k}{\alpha_k} \sqrt{\mathrm{Err}_y (\alpha_k, \gamma_k, k)}}_{(c)}
    \\ &
    +
    \underbrace{\frac{M_2}{\sqrt{K}} \sum_{k=1}^K k^{b-1} \sqrt{\mathrm{Err}_y (\alpha_k, \gamma_k, k)}}_{(d)}
    .
\end{split}
\end{equation}
From the first group $\sum_{k=1}^K (k^{a/2-1} + k^{b/2-1})$, we see that the error bound is minimized with small exponents $a$ and $b$.
For the following calculations, we repeatedly evaluate finite sums of the form $\sum_{k=1}^K k^{-s}$.
We bound these sums using piecewise functions $\mathcal{B}_K (s)$ defined as
\begin{equation}\label{eq:riemann_sum}
    \sum_{k=1}^K \frac{1}{k^s} \leq \mathcal{B}_K (s)
    \coloneqq
    \begin{cases}
        \zeta(s) & : s > 1 \\
        1 + \log K & : s = 1 \\
        1 + \frac{K^{1-s}}{1-s} & : s < 1 ,
    \end{cases}
\end{equation}
where $\zeta(s) = \sum_{k=1}^\infty k^{-s}$ is the Riemann zeta function for $s > 1$.
Furthermore, when evaluating terms where $s$ approaches 1 from above, we isolate the singular behavior using the expansion $\zeta(s) = 1/(s-1) + s_0 + \mathcal{O}(s-1)$, where $s_0$ is the Euler-Mascheroni constant.

\textbf{Term (a):}
From \Cref{lem:fast_mse}, we have up to multiplicative constants the bound
\begin{align*}
    &\frac{1}{\sqrt{K}} \sum_{k=1}^K k^{a-1} \sqrt{\mathrm{Err}_x (\alpha_k, \gamma_k, k)}
    \\ \leq&
    \frac{1}{\sqrt{K}} \sum_{k=1}^K k^{a-b/2-1} \left(1 + k^{(a-b)/2} + k^{(b-1)/2} \right)
    \\
    \leq& \frac{1}{\sqrt{K}} \left(\frac{K^{a-b/2}}{a-b/2}
    + \frac{K^{a - 1/2}}{a - 1/2}
    + \sum_{k=1}^K k^{3a/2 - b - 1}
    \right) .
\end{align*}
For $3a > 2b$, the bound becomes
\begin{align*}
    (a) \leq \frac{M_1}{\sqrt{K}} \left(\frac{K^{a-b/2}}{a-b/2} + \frac{K^{a-1/2}}{a-1/2} + \frac{K^{3a/2 - b}}{3a-2b} \right) .
\end{align*}

\noindent
\textbf{Term (b):}
Ignoring constant multiples,
\begin{align*}
    &\frac{1}{\sqrt{K}} \sum_{k=1}^K k^{a-1} \sqrt{\frac{\gamma_k}{\alpha_k} \mathrm{Err}_{x y}(\alpha_k, \gamma_k, k)}
    \\ \leq&
    \frac{1}{\sqrt{K}} \sum_{k=1}^K k^{a-1} \cdot \frac{\gamma_k}{\alpha_k} \left(k^{-1/2} + \alpha_k + \gamma_k + \sqrt{\gamma_k} \right)
\end{align*}
The first quantity is bounded by $\mathcal{B}_K (3/2 - 2a + b)$.
The second quantity is bounded by $\mathcal{B}_K (1 + b - a) / \sqrt{K}$, which necessitates $b > a$.
The third term is bounded by $\mathcal{B}_K (1 + 2 (b-a)) / \sqrt{K}$.
The last term behaves as $\mathcal{B}_K (1 - 2a + 3b/2)$.
We consider the regime $4a > 3b$.

Combining, we have that for $4a > 3b$, up to multiplicative constants
\begin{equation*}
\begin{split}
    (b) &\leq \frac{M_1}{\sqrt{K}}\Big(
    \mathcal{B}_K\left(\frac{1}{2}(3 - 2 (2a - b)) \right)
    \\ &
    +
    \zeta(1 + b-a) + \zeta(1 + 2 (b-a))
    +
    K^{2a - 3b/2}
    \Big)
    .
\end{split}
\end{equation*}

\noindent
\textbf{Term (c):}
Recall the error rate from \Cref{lem:slow_mse}:
\begin{align*}
    \sqrt{\mathrm{Err}_y (\alpha_k, \gamma_k, k)}
    \leq k^{-1/2} + k^{a/2-b}
    .
\end{align*}
Therefore, (c) is proportional to
\begin{align*}
    \frac{1}{\sqrt{K} } \sum_{k=1}^K k^{2a-b-1} \left(k^{-1/2} + k^{a/2-b}  \right) .
\end{align*}
The first quantity is bounded up to a constant multiple by $\mathcal{B}_K (3/2-2a+b)$.
The second quantity is bounded by a constant multiple of $\mathcal{B}_K (1 - 5a/2 + 2b)$.
Choosing $b < 5a/4$, we obtain that this term is $\mathcal{O}(K^{5a/2 - 2b})$.
Combining, we have
\begin{align*}
    (c) \leq \frac{M_1}{\sqrt{K}} \left(\mathcal{B}_K\left(\frac{3}{2} - 2a + b\right) + \frac{K^{5a/2-2b}}{5a-4b} \right).
\end{align*}
\textbf{Term (d):}
For this quantity, we again use the bound
\begin{align*}
    \sqrt{\mathrm{Err}_y (\alpha_k, \gamma_k, k)} \leq k^{-1/2} + k^{a/2 - b}
\end{align*}
to obtain
\begin{align*}
    (d) &= \frac{M_2}{\sqrt{K}} \sum_{k=1}^K k^{b-1} \sqrt{\mathrm{Err}_y (\alpha_k, \gamma_k, k)}
    \\ &\leq
    \frac{M_2}{\sqrt{K}} \sum_{k=1}^K \left(
        k^{b-3/2} + k^{a/2-1}
    \right)
    \\ &
    \leq \frac{M_2}{\sqrt{K}} \left(\frac{K^{b-1/2}}{b-1/2}
    + K^{a/2}
    \right) .
\end{align*}
Substituting all of (a)--(d) into \eqref{eq:W1_bound} and ignoring the multiplicative constants $M_1$ and $M_2$, we obtain the bound
\begin{align*}
    &\frac{1}{\sqrt{K}} \left(K^{a/2} + K^{b/2} \right)
    \\
    +& \frac{1}{\sqrt{K}} \left(\frac{K^{a-b/2}}{a-b/2} + \frac{K^{a-1/2}}{a-1/2} + \frac{K^{3a/2 - b}}{3a-2b}  \right)
    \\
    + & \frac{1}{\sqrt{K}} \Bigg(    \mathcal{B}_K\left(\frac{1}{2}(3 - 2 (2a - b)) \right)
    +
    \zeta(1 + b-a)
    \\ &+ \zeta(1 + 2 (b-a))
    +
    K^{2a - 3b/2}\Bigg)
    \\+ &
    \frac{1}{\sqrt{K}} \left(\mathcal{B}_K\left(\frac{3}{2} - 2a + b\right) + \frac{K^{5a/2-2b}}{5a-4b} \right)
    \\ &
    +\frac{1}{\sqrt{K}} \left(\frac{K^{b-1/2}}{b-1/2}
    + K^{a/2}
    \right)
\end{align*}
which is optimized over the step size exponents
\begin{equation}\label{eq:ab_constraint}
    1/2 < a < b < 2a - \frac{1}{2}
    .
\end{equation}
When $a$ and $b$ are chosen to satisfy \eqref{eq:ab_constraint}, we see that the dominant terms in the error bound reduces to
\begin{align*}
    \frac{1}{\sqrt{K}} \left( K^{b/2} +  \frac{K^{a-1/2}}{a-1/2}
    + \frac{1}{b-a} + \frac{K^{2a - b - 1/2}}{2a  - b - 1/2}
    \right)
    .
\end{align*}
The choice of time-scale gap $b - a$ that minimizes this upper bound is of the form $b - a = \Theta (1/\log K)$, which is subject to the constraint $ a < b < 2a - 1/2$.
If $b = 2a - 1/2$, then the bound is of the form
\begin{align*}
    \frac{1}{\sqrt{K}} \left( K^{b/2} +  \frac{K^{a-1/2}}{a-1/2}
    + \frac{1}{b-a} + \log K
    \right)         .
\end{align*}
Lastly if $b > 2a - 1/2$, the step sizes considered are subject to the constraint $b < 5a/4$ and the bound is of the form
\begin{align*}
    \frac{1}{\sqrt{K}} \left( K^{b/2} +  \frac{K^{a-1/2}}{a-1/2}
    + \frac{1}{b-a} + \zeta(3/2 - 2a + b)
    \right) .
\end{align*}
The constraint necessitates that when $a = 1/2 + c_a/\log K$ , then the time-scale gap must be sufficiently large: $b > 1/2 + 2 c_a/\log K$.
Increasing the choice of $b$ results in the $K^{b/2}$ term growing larger than $K^{1/4}$, and we conclude that the choice of step size optimizing our upper bound is of the form
\begin{align*}
    a = \frac{1}{2} + \frac{c_a}{\log K}, \quad b = \frac{1}{2} + \frac{c_b}{\log K} ,
    \quad 0 < c_a < c_b < 2c_a .
\end{align*}
\section{Comparison with a Baseline Algorithm}\label{sec:baseline}
It is instructive to compare the error bound of TSA-PR with that of a broader class of algorithms to demonstrate optimality.
Here, the error rate and asymptotic covariance of TSA-PR are compared with an oracle algorithm that has access to the system parameters, where it is shown that the two performance guarantees coincide.
Such an algorithm has been considered as a baseline in prior work \cite{polyakJuditsky,konda2004convergence,mokkadem2006convergence}.

Consider an oracle TSA algorithm that is allowed to design two gain matrices $Q_{ff}, Q_{ss}$ and estimate the solution $(x^*, y^*)$ with the sequence $\{(x_k^\dagger, y_k^\dagger)\}$ updated as
\begin{equation*}
    \begin{split}
        x_{k+1}^{\dagger} &= x_k^{\dagger} -  \alpha_k Q_{ff} \left(A_{ff} x_k^\dagger + A_{fs} y_k^\dagger - W_k\right) \\
        y_{k+1}^{\dagger} &= y_k^{\dagger} -  \gamma_k Q_{ss} \left(A_{sf} x_k^\dagger + A_{ss} y_k^\dagger - V_k \right) .
    \end{split}
\end{equation*}
A constraint is imposed on $Q_{ff}$ and $Q_{ss}$ so that the modified system with the gain matrix is asymptotically stable.
It was shown in \cite{mokkadem2006convergence} that if the step sizes are chosen as $\alpha_k \propto k^{-a}$ and $\gamma_k \propto 1/k$ with $a \in (1/2, 1)$, the slow-time-scale $\{y_k^\dagger\}_{k=1}^K$ converges in distribution to the solution $y^*$: $\sqrt{K} (y_K^\dagger - y^*) \to \mathcal{N}(0, \Sigma_{ss}^\dagger(Q_{ss}))$, where $\Sigma_{ss}^\dagger (Q_{ss})$ is the solution to the Lyapunov equation
\begin{align*}
    &\quad \left(Q_{ss} \Delta - \frac{1}{2} I \right) \Sigma_{ss}^\dagger (Q_{ss}) + \Sigma_{ss}^\dagger (Q_{ss}) \left(\Delta^T Q_{ss}^T - \frac{1}{2}I\right)
    \\ &= Q_{ss} \Delta \tilde{\Gamma}_{ss} \Delta^T Q_{ss}^T ,
\end{align*}
where $\tilde{\Gamma}$ is the block matrix
\begin{align*}
    \tilde{\Gamma} =\begin{pmatrix}
        \tilde{\Gamma}_{ff} & \tilde{\Gamma}_{fs} \\
        \tilde{\Gamma}_{sf} & \tilde{\Gamma}_{ss}
    \end{pmatrix} = A^{-1} \Gamma A^{-T} .
\end{align*}
A similar result holds for $x_k^\dagger$, where the other Schur complement $G$ appears in place of $\Delta$ in the above equation.
The minimum mean square error and the corresponding asymptotic covariance matrix is obtained by solving
\begin{align*}
    \min_{Q_{ff}} \mathrm{Tr} \Sigma_{ff}^\dagger (Q_{ff}) \eqcolon \mathrm{Tr} \Sigma_{ff}^*,
    \quad
    \min_{Q_{ss}} \mathrm{Tr} \Sigma_{ss}^\dagger(Q_{ss}) \eqcolon \mathrm{Tr} \Sigma_{ss}^* ,
\end{align*}
which can be shown to satisfy
\begin{equation*}
        \Sigma_{ff}^*
        = \tilde{\Gamma}_{ff},
        \qquad
        \Sigma_{ss}^*
        = \tilde{\Gamma}_{ss}
        .
\end{equation*}
This matches the asymptotic covariance achieved by TSA-PR, suggesting that TSA-PR is asymptotically optimal.
Our non-asymptotic analysis complements this perspective by providing insight into how quickly TSA-PR approaches its asymptotic performance.


\section{Lower Bounds}\label{sec:LowerBound}
First, we compare the asymptotic covariance of TSA-PR with the minimum variance achievable by unbiased estimators, i.e., the Cram\'{e}r-Rao lower bound (CRLB).
While not applicable in general, the lower bound serves as a comparison, illustrating that the covariance and error rate of TSA-PR matches that of the CRLB.
Next, we derive a lower bound on the expected error.
A comparison with Corollary \ref{cor:mae} demonstrates that PR averaging achieves the optimal $K^{-1/2}$ rate.
The dependence on problem-dependent constants is also shown to be tight when the condition number of the asymptotic covariance is small.

The lower bounds on the expected square error and the expected error are derived over any class of estimators that have access to the model parameters and observe noise explicitly.
The lower bounds will be described by considering the estimation problem
\begin{equation}\label{eq:linear_equation}
    A z^* = \mathbf{\vartheta},
\end{equation}
where an estimate $\hat{z}_K$ has access to the invertible matrix $A$ and to $K$ noisy observations $\mathbf{\vartheta} + N_k$ with centered noise sequence $\{N_k\}_{k=1}^K$.
The results below apply to both single- and two-time-scale algorithms, where $A$ is a block matrix for the latter and $z = (x, y)$.

\textbf{Asymptotic Covariance:}
A natural unbiased estimate $\hat{z}_K$ of the solution to \eqref{eq:linear_equation} is given by $\hat{z}_K = K^{-1} \sum_{k=1}^K A^{-1} (\mathbf{\vartheta} + N_k)$.
Therefore, it is common to compare the asymptotic covariance of the PR average $\bar{z}_K$ with a lower bound on the covariance of $\hat{z}_K$.
Stochastic approximation algorithms are unbiased when initialized with zero mean, $\mathbf{\vartheta} = 0$, and noise is centered.
In such cases, the smallest covariance of any unbiased estimate $\hat{z}_K$ of $z^*$ is given by the CRLB.
For $N \sim \mathcal{N}(0, \Gamma)$, in which case $\hat{z}_K \sim \mathcal{N}(0, \Sigma^* / K)$ with $\Sigma^* = A^{-1} \Gamma A^{-T}$, the lower bound is $\mathrm{Cov}(\hat{z}_K) \succeq \frac{1}{K} \Sigma^*$.
This lower bound is asymptotically tight for stochastic approximation algorithms as noted in \cite{polyakJuditsky}.

Next, we show using simple algebra that the asymptotic covariance $\bar{\Sigma}$ of TSA-PR matches the minimum covariance $\Sigma^*$ above.
With $z = (x, y)$ and $N = (W, V)$, the inverse $A^{-1}$ can be expressed using its Schur complements $G = A_{ff} - A_{fs} A_{ss}^{-1} A_{sf}$ and $\Delta$ as
\begin{align*}
    A^{-1} &= \begin{pmatrix}
        G^{-1} & - G^{-1} A_{fs} A_{ss}^{-1} \\
        - A_{ss}^{-1} A_{sf} G^{-1} & A_{ss}^{-1} (I + A_{sf} G^{-1} A_{fs} A_{ss}^{-1})
    \end{pmatrix}
    \\ & = \begin{pmatrix}
        A_{ff}^{-1}\left(I + A_{fs} \Delta^{-1} A_{sf} A_{ff}^{-1} \right) & - A_{ff}^{-1} A_{fs} \Delta^{-1} \\
        - \Delta^{-1} A_{sf} A_{ff}^{-1} & \Delta^{-1}
    \end{pmatrix} .
\end{align*}
Instantiating the lower bound with the above expression for $A^{-1}$, we obtain
\begin{align*}
    \mathrm{Cov} (\hat{z}_K)
    &\succeq
    \frac{1}{K}\mathrm{Cov}
    \begin{pmatrix}
        G^{-1}(W - A_{fs} A_{ss}^{-1} V)
        \\
        \Delta^{-1} (V - A_{sf} A_{ff}^{-1} W)
    \end{pmatrix},
\end{align*}
where we used the first and second expressions for $A^{-1}$ to write the upper and lower components of the right hand side.
This lower bound matches the asymptotic covariance of TSA-PR.

\textbf{Proof of Theorem \ref{thm:lower_bound}:}
We prove a more general statement which encompasses both the $d > 48 \log 2$ regime stated in \Cref{thm:lower_bound} as well as the low dimensional setting (all $d \geq 1)$.
We proceed by writing the minimization problem as a hypothesis test.
For this lower bound, we consider the general case where the right-hand side $\mathbf{\vartheta}$ is not necessarily zero.

The problem reduces to estimating the mean $z$ of a multivariate Gaussian distribution $\mathcal{N}(z, \Sigma)$ given $K$ i.i.d. observations, where $\Sigma = A^{-1} \Gamma A^{-T}$.
We derive minimax lower bounds using Fano's method (high-dimensional) and Le Cam's method (low-dimensional).

\textbf{High Dimensional Lower Bound:}
By the Gilbert-Varshamov bound \cite[Lemma 9.2.3]{duchi}, there exists a set of $M \geq \exp (d/8)$ hypotheses $\postrevision{\mathcal{Z}_K \coloneqq \{z_i\}_{i=1}^M}  \subset \{0, \delta\}^d$ such that $\postrevision{\lVert z_i - z_j\rVert_1} \geq d \delta/4$ for all $i \neq j$.
Let $p_i$ denote the distribution of the observations given parameter \postrevision{$z_i$}.
By Fano's inequality \cite{birge2005new,tsybakov}, the minimax risk is lower bounded by
\begin{align*}
    &\inf_{\hat{z}} \sup_{z \in \mathbb{R}^d}  \mathbb{E}\left[\lVert \hat{z} - z \rVert \right]
    \\ \geq &
    \delta \frac{\sqrt{d}}{4} \left(
        1 - \frac{\max_{i \neq j} \mathrm{KL}(p_i^K, p_j^K) + \log 2}{\log M}
    \right) .
\end{align*}
Using the tensorization property of the KL divergence for product measures $p_i^K$, we have $\mathrm{KL}(p_i^K, p_j^K) \leq (K/2) \lVert \Sigma^{-1} \rVert \delta^2 d$.
Substituting this into the bound with $\log M = d/8$ and choosing $\delta = (8 K \lVert \Sigma^{-1} \rVert)^{-1/2}$ yields
\begin{equation}\label{eq:high_dimension_lb2}
    \min_{\hat{z}_K} \max_z \mathbb{E}\left[\lVert \hat{z}_K - z \rVert\right]
    \geq \sqrt{\frac{d}{128 K} \frac{\lVert \Sigma \rVert}{\kappa(\Sigma)}}\left(\frac{1}{2} - 8 \frac{\log 2}{d}\right)
    .
\end{equation}
This quantity is non-negative when $d > 16 \log 2$; in particular, for $d > 48 \log 2$ dimensions the lower bound reduces to
\begin{equation*}
    \min_{\hat{z}} \max_{i \in [M]}  \mathbb{E}\left[\lVert \hat{z} - z_i \rVert \right]
    \geq
    \frac{1}{3} \sqrt{\frac{d}{128 K \lVert \Sigma^{-1} \rVert}} .
\end{equation*}
The statement expresses this lower bound in terms of $\lVert \Sigma \rVert$ and the condition number $\kappa(\Sigma)$.

\textbf{Low Dimensional Setting:}
The bound \eqref{eq:high_dimension_lb2} is vacuous for low dimensions.
For any $d \geq 1$, we prove
\begin{align*}
    \min_{\hat{z}_K} \max_{z^*} \mathbb{E}\left[\lVert \hat{z} - z^* \rVert\right]
    \geq
    \frac{1}{4 \sqrt{K \lambda_{\min}(\Sigma^{-1})}}
    .
\end{align*}
We apply Le Cam's method with two hypotheses
\begin{align*}
    p_0: z_0 \sim \mathcal{N}(0, \Sigma), \quad
    p_1: z_1 \sim \mathcal{N}(\delta, \Sigma) .
\end{align*}
The minimax risk is bounded using \cite[Theorem 2.2]{tsybakov}, where
\begin{align*}
    \inf_{\hat{z}} \sup_{z} \mathbb{E}\left[\lVert \hat{z} - z \rVert \right]
    \geq \frac{\lVert \delta\rVert }{2}  \left(1 - \mathrm{TV}(p_0, p_1) \right)
    .
\end{align*}
We now incorporate the product measures $p_0^K, p_1^K$ to account for i.i.d. observations.
The TV distance can be bounded above using Pinsker's inequality:
\begin{align*}
    \mathrm{TV}(p_0^K, p_1^K) &\leq \sqrt{\frac{K}{2} \mathrm{KL}(p_0, p_1)}
    = \sqrt{\frac{K}{4}\delta^T \Sigma^{-1} \delta } .
\end{align*}
Let $v$ be the eigenvector corresponding to the smallest eigenvalue of $\Sigma^{-1}$.
Setting $\delta = \mu v/\sqrt{K}$ for some $\mu > 0$ and $\lVert v \rVert = 1$ yields the lower bound on the minimax absolute error
\begin{align*}
    &\min_{\hat{z}_K} \max_{z^*} \mathbb{E}\left[\lVert \hat{z} - z^* \rVert \right]
    \\ &\geq
    \max_{\mu, \lVert v \rVert = 1} \frac{\mu}{2\sqrt{K}}
    \left(1 - \mu \sqrt{\frac{1}{4} v^T \Sigma^{-1} v }\right)
    \\ &\geq \max_{\mu > 0} \frac{\mu}{2 \sqrt{K}} \left(
        1 - \frac{\mu}{2} \sqrt{\lambda_{\min}(\Sigma^{-1}) }
    \right)
    .
\end{align*}
The maximizing lower bound is achieved when $\mu = 1/\sqrt{\lambda_{\min}(\Sigma^{-1})}$, which yields the lower bound
\begin{align*}
    \min_{\hat{z}_K} \max_{z^*} \mathbb{E}\left[\lVert \hat{z} - z^* \rVert\right]
    \geq
    \frac{1}{4 \sqrt{K \lambda_{\min}(\Sigma^{-1})}}
    .
\end{align*}

\bibliographystyle{IEEEtran}
\bibliography{IEEEabrv,references}

\end{document}